\documentclass[10pt]{article}
\usepackage[accepted]{tmlr}

\def\month{10}
\def\year{2026}
\def\openreview{\url{https://openreview.net/forum?id=N2iiNBZkVi}}

\usepackage[utf8]{inputenc}
\usepackage[T1]{fontenc}
\usepackage{hyperref}
\usepackage{url}
\usepackage{booktabs}
\usepackage{amsfonts}
\usepackage{amsmath}
\usepackage{nicefrac}
\usepackage{microtype}
\usepackage{xcolor}
\usepackage{graphicx}
\usepackage{multirow}
\usepackage{pifont}
\usepackage{algorithm}
\usepackage{algpseudocode}
\usepackage{booktabs}
\usepackage{makecell}
\newcommand{\gr}{\mathbf{g_r}}
\newcommand{\gf}{\mathbf{g_f}}
\newcommand{\grr}{\mathbf{g_r^R}}
\newcommand{\gfr}{\mathbf{g_f^R}}
\newcommand{\bgr}{\mathbf{\bar{g}_r}}
\newcommand{\bgrr}{\mathbf{\bar{g}_r^R}}
\newcommand{\bgf}{\mathbf{\bar{g}_f}}

\newcommand{\shat}{$\hat{\mathrm{S}}$}
\newcommand{\unoh}{UN$\hat{\mathrm{O}}$}
\newcommand{\unosh}{UN$\hat{\mathrm{O}}$-$\hat{\mathrm{S}}$}

\title{UNO: Unlearning via Orthogonalization in Generative Models}

\author{
  \name Pinak Mandal
  \email pinak.mandal@sydney.edu.au\\
  \addr University of Sydney, NSW 2006, Australia
  \AND
  \name Georg A.\ Gottwald
  \email georg.gottwald@sydney.edu.au\\
  \addr University of Sydney, NSW 2006, Australia
}

\begin{document}

\maketitle

\begin{abstract}
  As generative models become increasingly powerful and pervasive, the ability to unlearn specific data, whether due to privacy concerns, legal requirements, or the correction of harmful content, has become increasingly important. Unlike in conventional training, where data are accumulated and knowledge is reinforced, unlearning aims to selectively remove the influence of particular data points without costly retraining from scratch. To be effective and reliable, such algorithms need to achieve (i) forgetting of the undesired data, (ii) preservation of the quality of the generation, (iii) preservation of the influence of the desired training data on the model parameters, and (iv) a small number of training steps. We propose fast unlearning algorithms based on loss gradient orthogonalization for unconditional and conditional generative models. We consider the usual loss function of the original model restricted to the retain data and augment it by a term promoting the orthogonality between the two gradients of the loss function evaluated at the retain dataset and the forget dataset. We coin this method unlearning via orthogonalization (UNO) and show that the additional weak orthogonality constraint accelerates unlearning. The inclusion of the weak orthogonality constraint implies that each gradient descent step utilizes local information about the geometry via the Hessian of the loss function. Our geometry aware gradient descent step significantly accelerates the learning of the retain data loss function compared to a gradient descent which uses only gradient information. By alternating between an UNO and a gradient surgery step, a method we coin UNO-S, we are able to improve the overall execution time over UNO even further. We show that our algorithms are able to forget data while maintaining the fidelity of the original model. On standard image benchmarks, our algorithms achieve orders of magnitude faster unlearning times than their predecessors, such as gradient surgery. We demonstrate our algorithms with datasets of increasing complexity (MNIST, CelebA and ImageNet-1K) and for generative models of increasing complexity using VAEs and diffusion transformers.  
   \end{abstract}

\section{Introduction}\label{sec:intro}
Machine learning models are often trained on datasets that contain personal or sensitive information, such as medical records, financial data, or social media activity~\citep{mireshghallah2021privacy, truong2021privacy}. This reliance on personal data introduces substantial privacy risks, especially when models can unintentionally memorize or leak identifiable information; see~\citet{carlini2021extracting} for an in-depth exploration in the context of large language models (LLMs). Legal frameworks such as the General Data Protection Regulation (GDPR) and related EU laws have been established to address these issues~\citep{gdpr2016}. One of the central provisions is the right to be forgotten (RTBF), which grants individuals the ability to request the deletion of their personal data~\citep{kuner2020gdpr}. 
%It is increasingly likely that this obligation will become a standard requirement for machine learning services. 
Retraining large models from scratch each time such a request is received is computationally infeasible since the training costs are substantial~\citep{brown2020language, hoffmann2022training}. Machine unlearning refers to the removal of the influence of specific data points from a trained model without requiring full retraining. In the context of generative models {a particular form of unlearning is behavioural suppression. Behavioural suppression ensures that a model does not generate a specified unwanted output. This in itself does not imply that the data to be forgotten are not still encoded in the model parameters and can potentially still be recovered by adversarial attacks. Behavioural suppression} can be formalized as follows. Given a training dataset $\mathcal D=\mathcal D_r\sqcup \mathcal D_f$ partitioned into retain and forget datasets $\mathcal D_r$ and $\mathcal D_f$, respectively, and a model $\mathcal M_\theta$ trained on $\mathcal D$, the objective of unlearning is to update the model parameters $\theta$ in a way such that $P({\rm sim}(x, \mathcal D_f)\ge\delta)\le\varepsilon$ where $P$ denotes probability, $x$ is a sample generated by the updated model, ${\rm sim}$ is an appropriate similarity measure, and $\varepsilon, \delta$ are thresholds controlling the degree of forgetting. For an unlearning algorithm to be effective, it should (i) prevent the model from generating data resembling samples from $\mathcal{D}_f$, (ii) preserve the quality or fidelity of the generated samples, (iii) retain the influence of $\mathcal{D}_r$ on the model parameters, and (iv) require only a small number of training steps.

A simple approach to machine unlearning is to reverse the model update steps by performing gradient ascent on the loss function computed over the forget dataset $\mathcal{D}_f$. However, this method is susceptible to catastrophic forgetting, where the model loses knowledge far beyond just the targeted forget dataset~\citep{mccloskey1989catastrophic, luo2023empirical}. To mitigate this, several approaches combine gradient ascent on $\mathcal{D}_f$ with gradient descent on the retain dataset $\mathcal{D}_r$~\citep{yao2024machine}. The Gradient Difference (GDiff) method minimizes the difference of losses evaluated on the retain and forget datasets. Balancing the opposing updates in ascent-descent methods or weighing the loss terms properly in methods like GDiff is challenging since the forget and retain datasets might have significant size disparity, and the risk of catastrophic forgetting persists unless training hyperparameters such as the learning rate are finely tuned~\citep{bu2024unlearning}. Recently, multi-task optimization (MTO) techniques have inspired several unlearning algorithms~\citep{sener2018multi, yu2020gradient}. One such algorithm is gradient surgery~\citep{bae2023gradient} where gradient ascent is performed in a direction that is orthogonal to the loss gradient computed over the retain dataset. 
%While theoretically sound, 
This method, however, remains sensitive to the choice of hyperparameters and can suffer from catastrophic forgetting without careful tuning, see \ref{app:gs} for an example. \\

%\subsection*{Contributions}
{\bf{Contributions}}: In this work, we aim to advance the gradient surgery framework for unlearning in generative models. {Gradient surgery uses the loss function used to train the original model but restricted to the retain dataset. We modify this loss function by adding a term promoting the orthogonality between the gradients of the loss function evaluated at the retain dataset and the forget dataset. We coin this method unlearning via orthogonalization (UNO). We further augment UNO by an additional gradient surgery step, coined UNO-S which we will see improves the execution time over UNO. The decreased cosine similarity between the two loss gradients, caused by the weak orthogonality constraint, eliminates catastrophic forgetting. The weak orthogonality constraint implies that each gradient descent step involves the Hessian of the loss function. This additional information about the local geometry of the loss function is shown to significantly speed up the unlearning with a significant reduction in the number of steps required to achieve unlearning, compared to full retraining with gradient descent which only involves the information provided by the gradient. The inclusion of second-order derivatives in the loss function, however, does not prohibit the scalability of the method as we only require the application of the Hessian on the respective gradients which scales favourably \citep{pearlmutter1994fast} (see Section~\ref{sec:scalability}).}

%Like gradient surgery, 
Our proposed algorithms are general-purpose and {can be applied to different model architectures and datasets}. To demonstrate their effectiveness we will employ variational autoencoders (VAEs)~\citep{kingma2013auto, rezende2014stochastic} and diffusion transformers~\citep{peebles2023scalable} for three widely used benchmark image datasets MNIST~\citep{deng2012mnist}, CelebA~\citep{liu2015deep} and ImageNet-1K~\citep{deng2009imagenet}. 

{To demonstrate unlearning we restrict here to scenarios where the forget data consist of a specified class, such as forgetting a digit in MNIST, forgetting male faces or eyeglasses or nonsequential multiclass unlearning in CelebA, or replacement unlearning for conditional models where labrador retrievers are replaced by golden retrievers in ImageNet-1K. This allows us to verify the effectiveness of the respective unlearning methods with respect to behavioural suppression, for example, via counting the number of unwanted samples generated by the unlearned model. Behavioural suppression in itself does not exclude that the model still encodes information about the forget dataset. To investigate if our methods are capable of forgetting beyond behaviour suppression, we run loss-based membership inference attacks and relearning attacks on the unlearned models to see if unlearning exacerbates privacy risks and whether the forgotten data remain dormant inside the unlearned models. We find that our unlearning methods do not change the privacy risks of the original model under loss-based membership inference attacks. We find that for MNIST the unlearned model regains the lost knowledge slower than a retain-only retraining oracle suggesting that the forgotten data was indeed successfully unlearned by our model. However, this success is dataset dependent. For CelebA the unlearned model regains the lost knowledge faster than a retain-only retraining oracle indicating the forgotten data are still encoded in the unlearned model. 
%We do not address here the situation, common where a single data entry 
}

%% GAG: describe the class: feature sharing and not eftaure sharing. etc...

%% GAG: should we say soemthing about the MIA and relearning?

%\subsection*{Contributions}
%Our main contributions are as follows:
%\begin{enumerate}
%    \item We propose two new unlearning algorithms that regularize the main loss function with an additional term promoting orthogonality between loss gradients computed over the retain and forget datasets. We provide algorithms for both unconditional and conditional generative models. {Our methodology is applicable for general model architectures.}
    
%    \item We compare our algorithms against prior approaches, including gradient surgery and gradient ascent, evaluating both unlearning speed and the quality of generated samples. Our methods achieve orders of magnitude faster unlearning than gradient surgery, while retaining the influence of the desired training data unlike gradient ascent.  
%\end{enumerate}

%{
%\subsection*{Limitations}
{
{\bf{Limitations}}: In this work we are concerned with forgetting classes and assume that we have access to the forget and retain data. Our setting of working with retain and forget datasets, in principle, includes the situation of unlearning the influence of a singular data point on the parameters of a generative model. However, to evaluate the success of unlearning a single data point requires the computation of so-called influence functions which are hard to compute \citep{koh2017understanding}. Moreover, it is hard to assess the absence of a single data point in the generated data. We therefore limit our experiments here to forgetting classes with at least moderate cardinality (as few as $13$ images for ImageNet-1K). We do not, however, tackle the arguably harder problem of forgetting abstract concepts or reasoning as well as forgetting single sentences in language models. We further do not study the problem of continual application of forgetting updates which is common for machine learning service providers, as again it is difficult to measure the effectiveness of such continuous unlearning algorithms, both in terms of lack of standardized metrics and of computational costs.}

{The paper is organized as follows. In Section~\ref{sec:relatedwork} we provide a brief overview of related work. Section~\ref{sec:method} reviews gradient-based unlearning methods such as gradient-ascent and gradient surgery, and introduces an improved new version of gradient surgery as well as our new proposed gradient-orthogonality unlearning methods. Section~\ref{sec:res} explores the effectiveness of our unlearning methods for MNIST, CelebA and ImageNet-1K, and demonstrates the impact of the orthogonalization on the effectiveness and speed of unlearning and their scalability. We conclude in Section~\ref{sec:disc} with a discussion.}

Implementations of the proposed and baseline algorithms, together with the experimental data, are available at \href{https://github.com/pinakm9/forget}{https://github.com/pinakm9/forget}.

\section{Related work}
\label{sec:relatedwork}
 Early foundational work by~\citet{koh2017understanding} introduced influence functions as a principled approach for quantifying the impact of removing individual training points from machine learning models. The influence function is obtained from a higher-order Taylor expansion and involves the inverse of the Hessian of the loss function. Although influential, their technique is computationally demanding, limiting its scalability, particularly for large-scale neural networks~\citep{basu2020influence, guo2020fastif}. To address these computational challenges, recent studies have developed more efficient and scalable methodologies. For example, \citet{guo2020fastif} and \citet{schioppa2022scaling} proposed efficient approximations of influence functions that significantly reduce computational complexity. {Graph influence functions were introduced to achieve efficient unlearning in graph neural networks \citep{WuEtAl23,XiaoEtAl25}}. {\citet{ZhengEtAl26} achieve unlearning by replacing the inverse Hessian by a scalar.}  \citet{golatkar2020eternal} approximate the influence of individual data points on model parameters using the Fisher information matrix to approximate the inverse Hessian and use it to execute unlearning in deep networks. Building on this, Mixed-Privacy Forgetting~\citep{golatkar2021mixed} combines public and private data during training, enabling the selective removal of private data while preserving the utility of public data. Further, optimization-based frameworks such as SCRUB~\citep{kurmanji2023towards} approximate data removal for classification models such as ResNet~\citep{he2016deep} using a teacher-student formulation combined with checkpoint rewinding. \citet{ChenEtAl25} performed unlearning by optimizing a loss function that pushes predictions on the forget data towards an unseen third-party reference distribution while keeping the model close to its original behavior on the retain data. 

 %% GAG: we should somehow discuss how lots of papers do some higher-order Taylor expansion. This is where the Hessian comes in. I tried this. ok?
 
Unlearning in generative models introduces distinct challenges due to their capacity to implicitly memorize training data, complicating data removal without degrading generative quality. Addressing these, \citet{sun2025generative} introduced methods specifically designed to detect and mitigate unintended memorization in generative adversarial networks. \citet{selective_amnesia} proposed Selective Amnesia, which leverages continual learning frameworks to selectively remove specific concepts from deep generative models without compromising the overall data distribution learned by the model. \citet{moon2024feature} presented a method for unlearning features in VAEs and GANs by identifying unwanted features in the latent space. In the context of LLMs, recent works have tackled critical challenges such as selective forgetting of harmful or copyrighted content and aligning models to user preferences~\citep{jang2022knowledge, chen2023unlearn, qu2025frontier, pawelczyk2023context}. These methods employ parameter-efficient fine-tuning, low-rank adaptations, and in-context learning strategies to remove specific learned knowledge while minimally impacting overall model performance. Concept erasure or style unlearning has been recently studied in the context of text-to-image models, {for single-concept erasure} using negative classifier-free guidance ~\citep{gandikota2023erasing} {and for multiple-concept erasure using dynamic gradient masks~\citep{LiEtAl25}}, with remarkable results. \citet{neel2021descent} proposed Descent-to-Delete, which performs a short phase of projected gradient descent to approximate retraining without the undesired data. Negative Preference Optimization (NPO)~\citep{zhang2024negative} offers an alignment-inspired approach to machine unlearning by assigning lower preference or likelihood to data from the forget set. Through preference-based training, the model learns to reduce its reliance on forget data, often using pairwise comparisons or preference signals. Normalized Gradient Difference (NGDiff)~\citep{bu2024unlearning} approaches unlearning as a multi-task optimization problem, balancing the objectives of forgetting and retaining. By normalizing the gradient differences between these tasks and employing an adaptive learning rate scheduler, NGDiff provides stable training and effectively manages the trade-off between unlearning and model utility. \citet{cao2022machine} proposed a projection residual based method to remove the influence of undesired data. 

{Most of these unlearning methods are tailored for a specific model architecture. For example, the concept-erasure implementation proposed by \cite{gandikota2023erasing} is restricted to diffusion models and transformers, and the methods based on graph influence function by \citet{WuEtAl23,XiaoEtAl25} are restricted to graph neural networks. An attractive framework that works for all model architectures is provided by combining variants of gradient ascent and descent steps along the gradients evaluated on the forget and retain datasets, respectively.} Gradient surgery~\citep{bae2023gradient} attempts to maximize the loss function in a direction orthogonal to the loss gradient evaluated on the retain dataset. While promising for generative models, gradient surgery can suffer from inefficiency when there is significant overlap between loss gradients computed on the retain and forget data, and may even cause catastrophic forgetting. MUNBa~\citep{wu2025munba} addressed such gradient conflicts within a two-player cooperative game framework based on Nash bargaining. SalUn~\citep{fan2024salun} identified parameters most relevant to the forget set using gradient-based saliency and selectively updates them to remove the targeted information. We aim to improve upon gradient surgery by explicitly enforcing orthogonality between conflicting gradients. Our algorithms exhibit no catastrophic forgetting, achieve fast unlearning speeds, and are robust to hyperparameter selection.

For a comprehensive overview of unlearning techniques for large language models, including method categorization and scale-specific challenges, see \citet{blanco2025digital}.
For a broad taxonomy of machine unlearning across centralized, distributed, and privacy-critical settings with a focus on open problems and verification, see \citet{wang2024machine}.

\section{Unlearning via orthogonalization}
\label{sec:method}
We now describe the unlearning algorithms used to produce the results presented in this paper. The pseudocode for all algorithms presented in this section can be found in Appendix~\ref{app:algo}. We first describe classical gradient ascent and gradient surgery before introducing our Unlearning via Orthogonalization (UNO) algorithms with and without surgery.

\subsection{Gradient ascent}
We begin by introducing the simplest approach, namely, gradient ascent. Given a pretrained model $\mathcal{M}_\theta$ with trained parameters $\theta=\theta^\star$, trained using a loss function $\mathcal{L}(\mathcal{M}_\theta,x)$ on a dataset $\mathcal{D}$, unlearning can be induced by maximizing the loss function on the forget data, which can be done with the update step:
\begin{align}
    \theta_{k+1} = \theta_k + \eta\, \gf,\tag{A}\label{eq:A}
\end{align}
where $\theta_k$ represents the model parameters after the $k$-th training step with $\theta_0=\theta^\star$, $\eta$ is the learning rate, and $\gf$ is the gradient of the loss function evaluated over the forget data (we omit the index of $\theta$ in the definition below for brevity),
\begin{align}
    \gf = \frac{1}{|\mathcal D_f|}\sum_{x\in \mathcal D_f}\nabla_\theta\mathcal L(\mathcal M_\theta, x).\label{eq:gf}
\end{align}
This approach, however, may delete knowledge acquired on the retain data $\mathcal D_r$ if $\gf$ significantly projects onto $\gr$, the gradient of the loss function evaluated over the retain data,
\begin{align}
    \gr = \frac{1}{|\mathcal D_r|}\sum_{x\in \mathcal D_r}\nabla_\theta\mathcal L(\mathcal M_\theta, x).\label{eq:gr}
\end{align}
A naive way to prevent the model from forgetting retain data is to perform alternating ascent in the direction of $\gf$ and descent in the direction of $\gr$:
\begin{align}
    \theta_{k+1} =\begin{cases}&\theta_k + \eta\,\gf,\quad\text{if } k\text{ is even},\\
        &\theta_k - \eta\,\gr,\quad\text{if } k\text{ is odd}.
    \end{cases} \tag{A-D}\label{eq:A-D}
\end{align}

This simple modification, which we will refer to as ascent-descent, does not safeguard against catastrophic forgetting, as we will see in Section~\ref{sec:res}. See Appendix~\ref{app:catastrophy} for examples of catastrophic forgetting induced by gradient ascent and ascent-descent.

\subsection{Gradient surgery}
Similar challenges also arise in a related subfield of machine learning: multi-task optimization where a model must learn to perform new tasks without compromising performance on earlier tasks~\citep{crawshaw2020multi}. If the loss gradient corresponding to the new task points in a direction opposing the loss gradients corresponding to the old tasks, the model risks losing its previously learned skills with each new gradient descent step, paralleling catastrophic forgetting.  In multi-task optimization, gradient surgery refers to techniques that modify task-specific gradients during training to reduce such interference between tasks. When gradients from different tasks conflict, i.e., point in opposing directions, methods like PCGrad project gradients to minimize this conflict, allowing the model to learn multiple tasks more effectively without one task hindering the progress of another~\citep{yu2020gradient}. 

Gradient surgery can be used to reduce the potential conflict between $\gf$ and $\gr$ to improve the vanilla gradient ascent~\citep{bae2023gradient} by explicitly removing the orthogonal projection of $\gr$ from $\gf$ before taking the ascent step:  
\begin{equation}
\begin{aligned}
    &\bgf = \gf - \frac{\gr\cdot\gf}{\gr\cdot\gr}\gr,\\
    & \theta_{k+1} = \theta_k + \eta \, \bgf.
\end{aligned}\tag{SA}\label{eq:SA}
\end{equation}
While this modified ascent reduces over-unlearning compared to vanilla ascent, it does not fully resolve the issue, and still requires careful tuning of $\eta$ to avoid catastrophic forgetting. Therefore, we introduce another version of gradient surgery which we find to be more stable.
%and use it throughout, for generating the results in Section~\ref{sec:res}. 

Rather than performing ascent along a modified $\gf$-direction, we propose to perform descent along a modified $\gr$-direction resulting in the following update:
\begin{equation}
\begin{aligned}
    &\bgr = \gr - \frac{\gr\cdot\gf}{\gf\cdot\gf}\gf,\\
    & \theta_{k+1} = \theta_k - \eta \,\bgr ,
\end{aligned}\tag{S}\label{eq:S}
\end{equation}
which aims at minimizing the loss function in directions orthogonal to $\gf$. %Outside of the context of unlearning, such surgery 

{In the original formulation (SA) by \citet{bae2023gradient} the ascent step aims at forgetting without explicitly trying to remember the retain data. In contrast, our modification (S) performs a descent step that forgets by only trying to remember the retain data.} 
%In other words (SA) tries to unlearn the forget data by deleting information about all training data, whereas (S) tries to unlearn the forget data by only trying to remember the retain data.
Indeed, as we will show, the modification (S) of gradient surgery does not suffer from catastrophic forgetting, is robust to the choice of $\eta$, and consequently achieves faster unlearning speeds compared to \eqref{eq:SA} with larger values of $\eta$. For a comparison of these two versions of gradient surgery: \eqref{eq:SA} and \eqref{eq:S}, see Appendix~\ref{app:gs}.

{We remark that (S) and (SA) modify the gradient of the loss function but neither method can be written as a gradient descent method on a modified loss function.} 

\subsection{UNO and UNO-S}
In the scenario when $\gf$ is orthogonal to $\gr$, \eqref{eq:SA} is equivalent to gradient ascent \eqref{eq:A} without the risk of losing desired knowledge. Furthermore, \eqref{eq:S} is equivalent to retraining the model on the retain data, without the risk of relearning all or parts of the forget data. Therefore, we propose a modified loss function that attempts to enforce this ideal scenario with the help of an orthogonality {promoting} regularization term,
\begin{align}
    \mathcal{L}_{\rm UNO} &= \frac{1}{|\mathcal D_r|}\sum_{x\in\mathcal D_r}\mathcal{L}(\mathcal{M}_\theta, x) +  \beta_o \left(\frac{\gr \cdot \gf}{\|\gr\|\|\gf\|}\right)^2 \nonumber \\
    &{= \frac{1}{|\mathcal D_r|}\sum_{x\in\mathcal D_r}\mathcal{L}(\mathcal{M}_\theta, x) +  \beta_o \, \cos^2 \vartheta}
    , 
    \label{eq:loss-uno}
\end{align}
where $\beta_o$ is a regularization parameter {and we introduced the cosine similarity
\begin{align}
\cos \vartheta = \frac{\gr \cdot \gf}{\|\gr\|\|\gf\|}.
\label{eq:cossim}
\end{align}
}
Note that we only use the retain data to construct the first term in \eqref{eq:loss-uno} to mimic the ideal retraining scenario mentioned above.

Unlearning via orthogonalization (UNO) is achieved by performing gradient descent on this modified loss function with
{
\begin{align}
    \theta_{k+1} &= \theta_k - \eta \, \nabla_{\theta_k}\mathcal L_{\rm UNO}\tag{UNO}\label{eq:UNO}\\
&= \theta_k - \eta \, \gr + 2 \beta_o \eta \, \sin \vartheta \cos \vartheta \, \nabla \vartheta. 
\end{align}
The additional term points in the direction in parameter space in which the angle $\vartheta$ between the retain and forget gradients changes most rapidly, or more concisely it points in the direction of steepest increase of their cosine similarity. The update step can be written explicitly in terms of second-order derivatives of the loss function as
\begin{align}
    \theta_{k+1} &= \theta_k - \eta \, \gr 
    -2 \beta_o \eta\, \frac{\gr \cdot \gf}{\lVert \gr\rVert^2 \lVert \gf\rVert^2}
    \left[
   {\rm{H}}_r \bgf     
    +
    {\rm H}_f \bgr
    \right] , 
    \label{eq:hess-update}
\end{align}
where 
\begin{align}
{\rm H_r} = \frac{1}{|\mathcal D_r|}\sum_{x\in \mathcal D_r}\nabla_\theta^2\mathcal L(\mathcal M_\theta, x) \qquad {\rm{and}} \qquad  
{\rm{H}}_f = \frac{1}{|\mathcal D_r|}\sum_{x\in \mathcal D_f}\nabla_\theta^2\mathcal L(\mathcal M_\theta, x)
\label{eq:Hess_rf}
\end{align}
are the Hessians over the retain and forget data, respectively. In practice, we evaluate the full update with automatic differentiation. For $\beta_o=0$, UNO reduces to retraining on only the retain data, which practitioners avoid performing as it requires too many update steps.} To gain intuition on the effect of orthogonalization, we present in Appendix~\ref{app:lin} an analysis for a linear model in a discriminative context.\\ 
%since it would be wasteful to minimize loss evaluated on the forget data. 

We further propose a hybrid algorithm that applies the \eqref{eq:UNO} update step and the \eqref{eq:S} update step alternately which we refer to as UNO-S:
\begin{align}
    \theta_{k+1} =\begin{cases}&\theta_k - \eta\nabla_{\theta_k}\mathcal L_{\rm UNO},\quad\text{if } k\text{ is even},\\
        &\theta_k - \eta \, \bgr,\qquad\qquad\;\text{if } k\text{ is odd}.
    \end{cases} \tag{UNO-S}\label{eq:UNO-S}
\end{align}
The rationale behind this 2-step update is that the UNO update step attempts to enforce orthogonality between $\gf$ and $\gr$, which helps the subsequent surgery step to effectively resolve the conflict between them. 
%{We will see that UNO-S is capable of replacement unlearning where the forget and the retain set do not share many features, whereas UNO requires the two datasets to share features.}  

\subsection{Replacement unlearning for conditional generative models}\label{ssec:replace}
In the case of conditional generation, we can aim to replace the generation corresponding to an undesired condition $c_f$ with the generation corresponding to a target condition $c_t$. In this scenario it is natural to minimize the following quantity,
\begin{align}
    \mathcal L^{R}=\frac{1}{|\mathcal D_t|}\sum_{x\in\mathcal D_t}\mathcal{L}(\mathcal M_\theta(c_f), x
    ) + \frac{1}{|\mathcal D_t|}\sum_{x\in\mathcal D_t}\mathcal{L}(\mathcal M_\theta(c_t), x
    ),
\end{align}
where $\mathcal D_t$ is the data corresponding to $c_t$. The first term enforces replacement and the second term represents the conditional variant of the retain loss function that appears in \eqref{eq:gr}. Using $\mathcal L^R$, we can devise the replacement unlearning variants of \eqref{eq:S}, \eqref{eq:UNO} and \eqref{eq:UNO-S} by defining
\begin{align}
\mathcal{L}_{\rm UNO}^R = \mathcal L^R +  \beta_o \left(\frac{\grr \cdot \gfr}{\|\grr\|\|\gfr\|}\right)^2,\label{eq:loss-replace-uno}
\end{align}
where
\begin{align}
&\mathbf{g_r^R} = \nabla_\theta\mathcal L^R,\label{eq:grr}\\
&\mathbf{g_f^R} = \frac{1}{|\mathcal D_f|}\sum_{x\in\mathcal D_f}\nabla_\theta\mathcal{L}(\mathcal M_\theta(c_f), x).
\label{eq:gfr}
\end{align}
Here $\mathcal D_f$ again denotes the forget data associated with the condition $c_f$. Performing gradient descent on \eqref{eq:loss-replace-uno} yields the replacement unlearning variant of UNO.
 For the surgery update step \eqref{eq:S}, we replace $\bgr$ with 
\begin{align}
\bgrr = \grr - \frac{\grr\cdot\gfr}{\gfr\cdot\gfr}\gfr.
\label{eq:bgrr}
\end{align}
Alternating between the replacement unlearning variants of UNO and gradient surgery yields the replacement unlearning variant of UNO-S.

\subsection{Classifier-assisted unlearning}\label{ssec:hat}
We further consider the case when we may have access to a binary classifier that distinguishes forget data from retain data and we can leverage this extra information to accelerate unlearning algorithms. We can use this classifier to identify every sample generated by our model as either a retain or forget sample, and compute the probability $p_r$ that a generated sample is a retain sample. This associates our generative model with a Bernoulli distribution with probability of success $p_r$. We would like this distribution to have probability of success close to %$1$ with 
$1-\alpha$ where $\alpha$ is a small positive threshold controlling the degree of forgetting. We can promote this by adding the following term to our loss function,

\begin{align}
     \beta_h d_{\rm KL}=\beta_h \left[p_r \log\left(\frac{p_r}{1 - \alpha}\right) + (1 - p_r) \log\left(\frac{1 - p_r}{\alpha}\right)\right], \label{eq:loss-h}
\end{align}
where $\beta_h$ is a regularization parameter, and $d_{\rm KL}$ represents the KL divergence between the computed and the desired Bernoulli distributions. Small positive values of $\alpha$ ensure stable computation of the KL divergence. Recalling that $p_r$ is a function of the model and its parameters, we can now use the modified loss function in place of the original loss function in the previously described algorithms. We use the hat symbol ( $\hat{}$ ) to denote unlearning algorithms that operate with the additional loss term~\eqref{eq:loss-h}. For example, gradient surgery (S), UNO, and UNO-S become \shat, \unoh, and \unosh, respectively, when~\eqref{eq:loss-h} is included. Addition of the new term yields the following modified definitions of $\gf$ and $\gr$:
\begin{align}
    &\gf = \frac{1}{|\mathcal D_f|}\sum_{x\in \mathcal D_f}\nabla_\theta\mathcal L(\mathcal M_\theta, x) + \beta_h \nabla_\theta d_{\rm KL},\label{eq:gfh}\\
    &\gr = \frac{1}{|\mathcal D_r|}\sum_{x\in \mathcal D_r}\nabla_\theta\mathcal L(\mathcal M_\theta, x) + \beta_h \nabla_\theta d_{\rm KL}.\label{eq:grh}
\end{align}
Using \eqref{eq:gfh}, \eqref{eq:grh} with \eqref{eq:S} gives us \shat. Similarly, the update rule for \unoh~can be written as,
\begin{equation}
\begin{aligned}
    &\mathcal{L}_{\rm UN\hat{O}} = \frac{1}{|\mathcal D_r|}\sum_{x\in\mathcal D_r}\mathcal{L}(\mathcal{M}_\theta, x) +  \beta_o \left(\frac{\gr \cdot \gf}{\|\gr\|\|\gf\|}\right)^2 + \beta_h d_{\rm KL}, 
    \\
    &\theta_{k+1} = \theta_k - \eta \, \nabla_{\theta_k}\mathcal L_{\rm UN\hat{O}}.
\end{aligned}\tag{\unoh} \label{eq:UNOh}
\end{equation}
Alternating update steps of \unoh~and \shat~gives us \unosh. Since the KL divergence term promotes unlearning of the forget data by preventing generation of forget samples, we also test the following update rule which is equivalent to \unoh~with $\beta_o=0$,
\begin{equation}
\begin{aligned}
    &\mathcal{L}_{H} = \frac{1}{|\mathcal D_r|}\sum_{x\in\mathcal D_r}\mathcal{L}(\mathcal{M}_\theta, x) + \beta_h d_{\rm KL}, 
    \\
    &\theta_{k+1} = \theta_k - \eta \, \nabla_{\theta_k}\mathcal L_{H}.
\end{aligned}\tag{H} \label{eq:H}
\end{equation}
We call the resulting unlearning algorithm histogram unlearning and denote it by H. Appendix~\ref{app:un-w-cls} reports results for classifier-assisted unlearning on MNIST and CelebA.

\section{Results}\label{sec:res}
To test how far the additional orthogonalization of the gradients of the loss function evaluated on the retain and forget data provided by UNO and UNO-S and their conditional variants improves upon standard unlearning methods with respect to time to unlearn and accuracy, we test our algorithms on several datasets and several pretrained models. In particular, we evaluate the algorithms introduced in Section~\ref{sec:method} on VAEs trained on MNIST~\citep{deng2012mnist} and CelebA~\citep{liu2015deep}, as well as on diffusion transformers trained on ImageNet-1K~\citep{deng2009imagenet}. For VAEs, we compare our methods against standard gradient ascent, gradient surgery, and the feature unlearning method of \citet{moon2024feature}. For diffusion models, we compare them against the concept erasure method of \citet{gandikota2023erasing}. Each algorithm was tested $10$ times to generate statistics. For the training losses used to train the original VAEs, training data, experiment hyperparameters, and model sizes, refer to Appendix~\ref{app:setup}. The architecture of the models can be found in the code provided in Section~\ref{sec:intro}. All experiments were done on an A100 GPU provided by Google Colab. Table~\ref{tab:param} in Appendix~\ref{app:setup} documents the hyperparameters used in our experiments.

\subsection{Performance metrics}
In order to assess the speed of unlearning we use classifiers trained on the datasets and track the fraction of generated samples that are classified as forget samples after each model update or training step. We refer to the fraction of forget samples in generated data as the f-ratio. We define the \textbf{time to unlearn} as the execution time of the unlearning algorithm until the f-ratio drops below a chosen threshold $\tau$. On MNIST and CelebA, our classifiers achieve $\sim 98\%$ top-1 accuracy, whereas on ImageNet-1K our classifier achieves $\sim 82\%$ top-1 accuracy. Accordingly, we set $\tau = 0.02$ for MNIST and CelebA, and $\tau = 0.18$ for ImageNet-1K. We evaluate the quality of the generated images by computing the \textbf{Fréchet Inception Distance (FID)}. We also report the execution \textbf{time per training step} ($\Delta t$), however, we remark that a larger time per step does not necessarily indicate slower unlearning, and vice versa. Lastly, we report the \textbf{$5$-step f-ratio}, defined as the fraction of images in the generation that are detected as belonging to the class, or as having the feature that we are trying to forget, after five steps of the unlearning algorithm.

\subsection{MNIST}
\begin{figure}[h]
    \centering
    \includegraphics[width=0.6\linewidth]{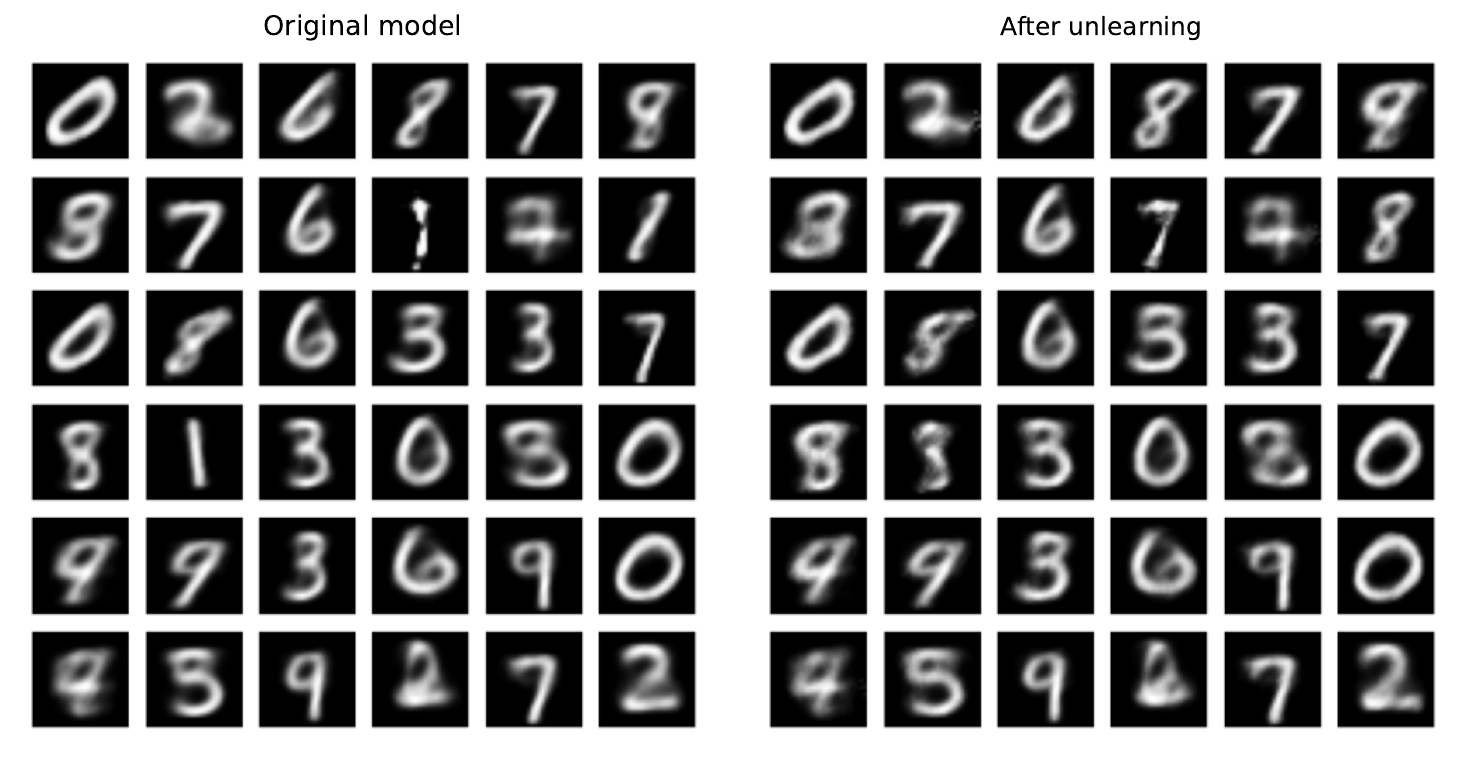}
    \caption{MNIST samples generated by the original model (left) and after unlearning digit "$1$" with UNO (right), using identical noise inputs for the decoder.}
    \label{fig:mnist-transform}
\end{figure}

\begin{figure}[h]
    \centering
    \includegraphics[width=0.6\linewidth]{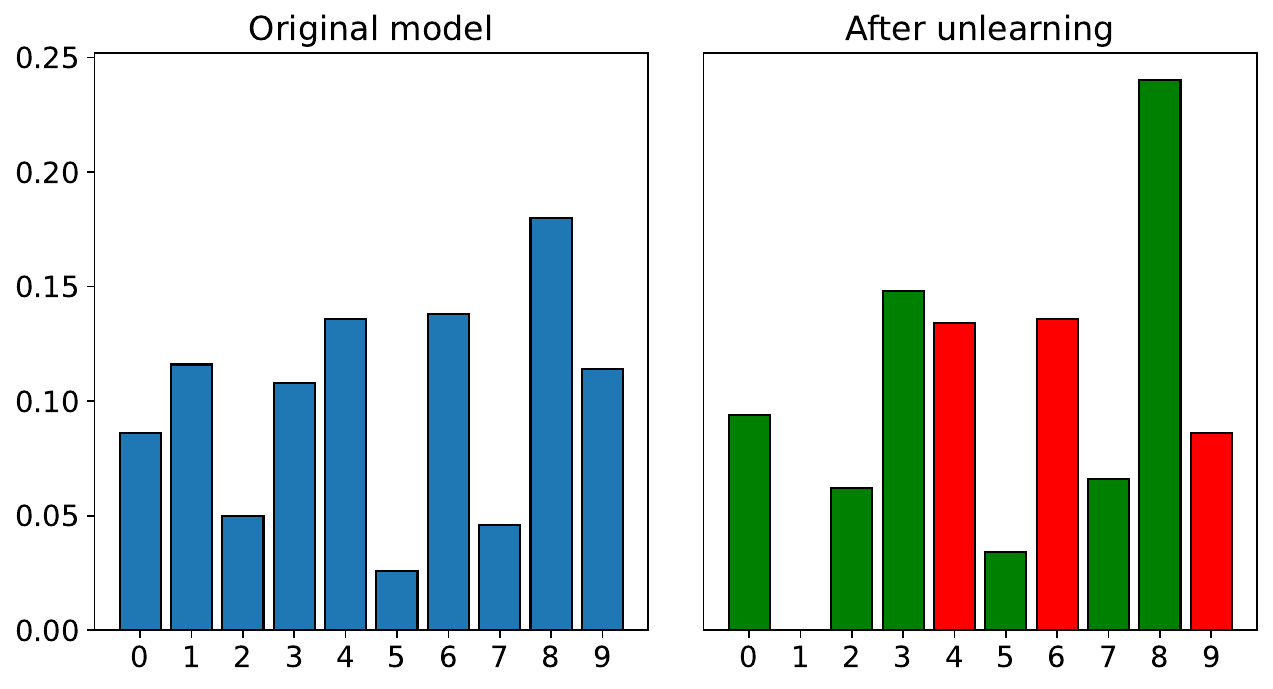}
    \caption{Distribution of generated digits before (left) and after unlearning (right), for a single run of UNO. Each histogram shows data for $500$ generated samples. A bar in the right panel is colored green if the fraction of the corresponding digit increases after unlearning, and red if it decreases.}
    \label{fig:mnist-dist}
\end{figure}

\begin{figure}
    \centering
    \includegraphics[width=0.5\linewidth]{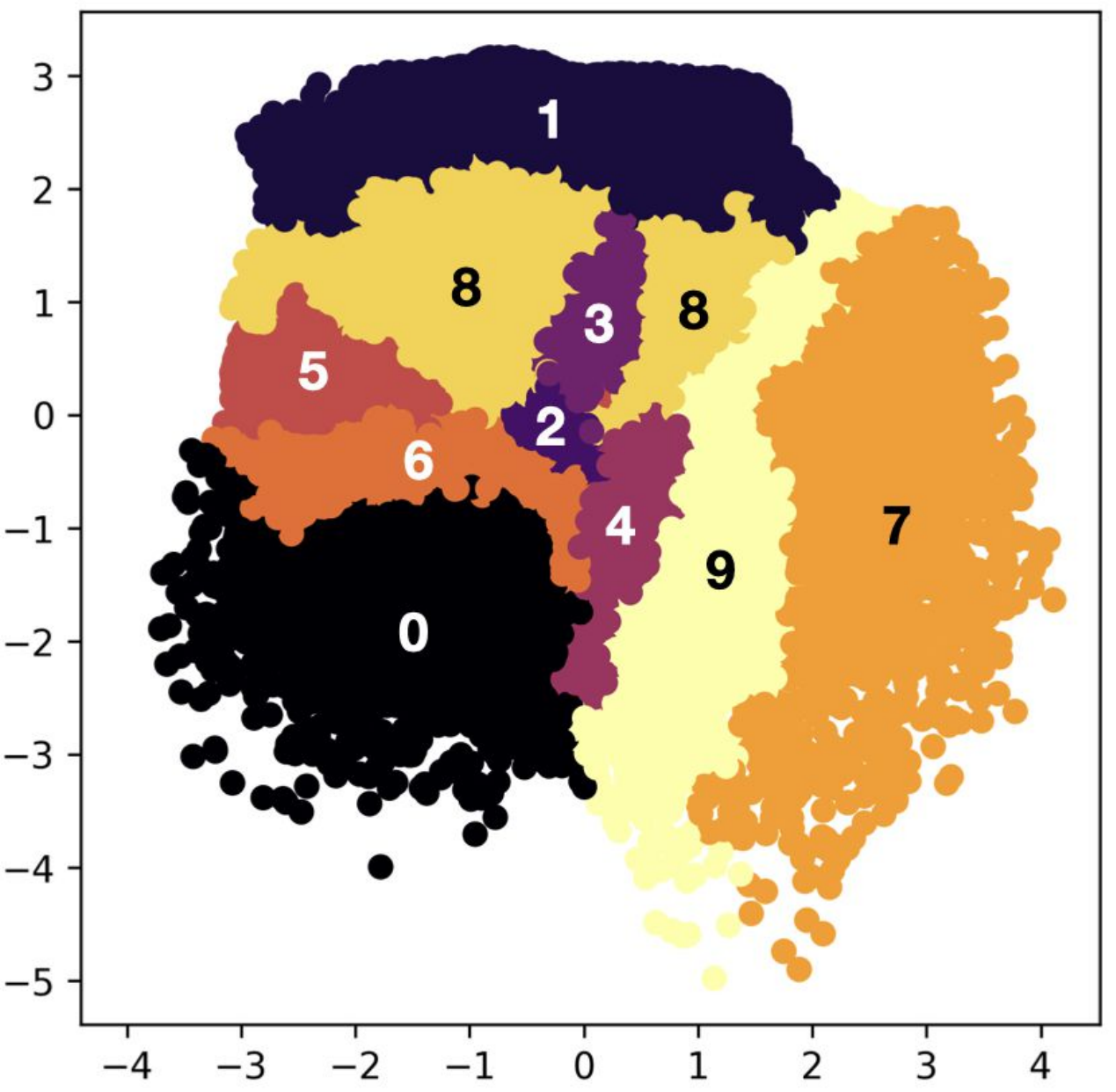}
    \caption{Distribution of digits in the latent space of the original model for MNIST. 
    %Noise sampled with mean and variance restricted to each coloured region generates predominantly the indicated digits.}
    The colors are obtained by mapping $50,000$ images to the two-dimensional latent space, labelling them according to the color associated with the image. To obtain smooth regions we performed averaging over nearby points in latent space.} 
   %most frequently generated digit from noise sampled from that box.}
    %We divide the two-dimensional latent space into regions; for each region, we sample noise from it, generate digits, and color the region by whichever digit appears most often.  }
    \label{fig:cartography}
\end{figure}

We use a $0.6$M-parameter VAE with a $2$-dimensional latent space, trained for $200$ epochs on $60,000$ images, as our original model. We attempt to unlearn the digit "$1$" by running the algorithms for $530$ training steps with a mini-batch size of $128$, and a learning rate of $\eta=10^{-3}$. Figure~\ref{fig:mnist-transform} shows samples generated before and after unlearning with UNO, using the same noise samples for ease of comparison. The $1$'s in the original generation (left) transform into $7, 8$ and $3$ after unlearning (right). The non-$1$ digits remain nearly unchanged. Even though $1$'s can transform into different digits, they more frequently turn into $8$'s, followed by $3$'s, as seen in Figure~\ref{fig:mnist-dist}. To understand this, we color regions in the latent space, which is two-dimensional in our case, according to the most frequently produced digits. Figure~\ref{fig:cartography} shows the distribution of digits in the latent space for the original model. We clearly see that the region corresponding to the digit $1$ shares the largest border with the region corresponding to digit $8$. This proximity in latent space makes it easier for the unlearning algorithms to transfer the probability mass of $1$ to that of $8$. Figure~\ref{fig:cartography} suggests that forgetting $7$ for our original model should mostly increase the frequency of $9$, which we have experimentally verified. If the goal is uniform generation across the retain classes, one may utilize a Kullback-Leibler divergence loss term promoting uniformity, assuming the availability of a classifier for all classes.

%Figure~\ref{fig:mnist} and 
%To assess the performance of our methods we compare with several unlearning methods. In particular, we compare with gradient ascent, ascent-descent, gradient surgery and feature unlearning for VAEs by Moon et al.~\citet{moon2024feature}, and concept erasure for diffusion models by Gandikota et al.~\citet{gandikota2023erasing}. 

We report in Table~\ref{tab:res} on the FID, the total time to unlearn, the time $\Delta t$ to execute a single step and the 5-step f-ratio for our method and for gradient ascent, ascent descent, gradient surgery and feature unlearning for VAEs by \citet{moon2024feature}. We provide pseudocode for all methods in Appendix~\ref{app:algo}. Table~\ref{tab:res} shows that UNO-S achieves the fastest unlearning time, closely followed by UNO. Both methods have similar fidelity as the original model, indicated by the FID. Gradient ascent, while fast at unlearning, suffers from catastrophic forgetting, resulting in a large FID. Ascent-descent also experiences catastrophic forgetting and is significantly slower at unlearning than gradient ascent. See Appendix~\ref{app:catastrophy} for examples of such catastrophic forgetting induced by gradient ascent and ascent-descent. Gradient surgery, while preserving image quality, is $\sim20$ times slower than UNO and UNO-S at unlearning. Even though UNO takes $\sim3$ times longer to execute a training step compared to gradient surgery, it still achieves orders of magnitude faster unlearning speed and has the smallest 5-step f-ratio implying that it requires fewer iterations to forget. Since one step of surgery is faster than one step of UNO, UNO-S overall is slightly faster than UNO, as the time per training step is roughly averaged over the two algorithms. 

% \begin{figure}[!htp]
%     \centering
% \includegraphics[width=\linewidth]{plots/mnist.pdf}
%     \caption{Time to unlearn (left) and FID (right) for various unlearning algorithms applied on MNIST. The middle panel shows the fraction of generated images classified as "1", averaged over $10$ runs, as a function of training steps. Standard deviations over $10$ independent runs are shown as one-sided error bars in the left and right panels. For time to unlearn and FID, the rank of each algorithm is also indicated below its name, with $1$ being the best.}
%     \label{fig:mnist}
% \end{figure}

\subsection{CelebA}
\begin{figure}[h]
    \centering
\includegraphics[width=\linewidth]{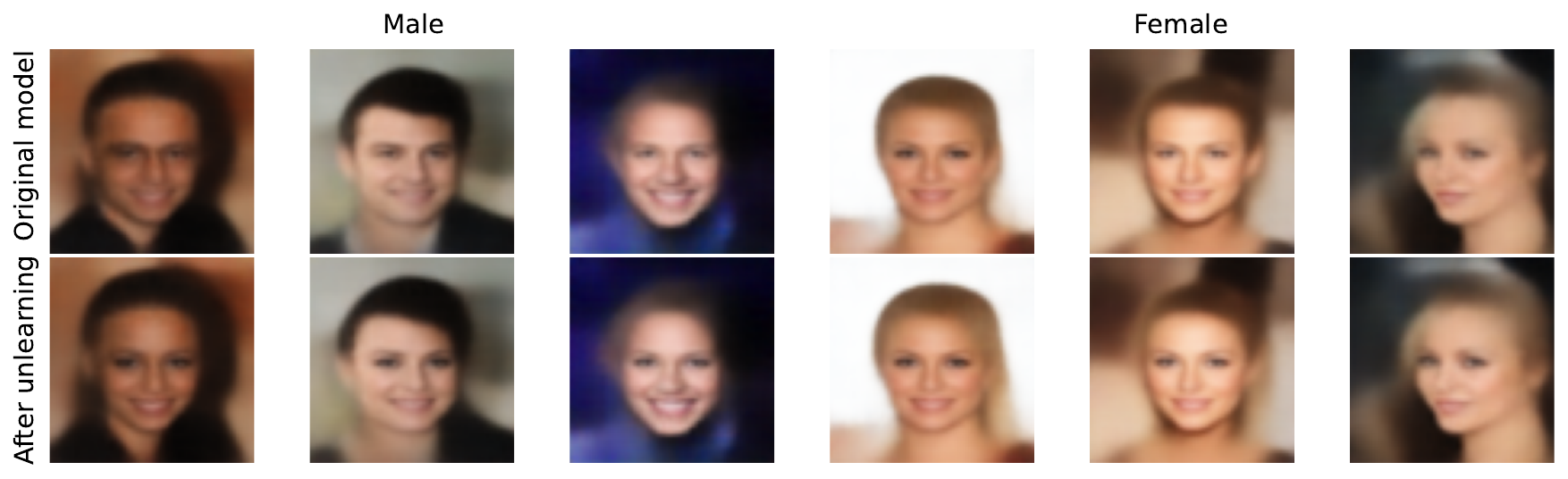}
    \caption{CelebA samples generated by the original model (top) and after unlearning "male" faces with UNO (bottom), using identical noise inputs for the decoder.}
    \label{fig:celeba-transform}
\end{figure}

We use an $8.7$M-parameter VAE with a $512$-dimensional latent space, trained for $200$ epochs on $202,599$ images at $64 \times 64$ resolution, downsampled from the original $178 \times 178$ resolution, as our original model. The dataset classifies each image as either female or male. We attempt to unlearn "male" faces by running the algorithms for $659$ training steps with a mini-batch size of $128$, and a learning rate of $\eta=10^{-3}$. Approximately $29\%$ of the faces generated by the original model are male. Figure~\ref{fig:celeba-transform} shows samples generated before and after unlearning with UNO, using the same noise samples in the decoder. We observe that images that are classified as male are successfully converted into images that are classified as female, and that feminine features are enhanced after unlearning, even when the originally generated face was already classified as female. The original image remains nearly unchanged if it contains few or no male-specific features; see, for example, the last pair from the left in Figure~\ref{fig:celeba-transform}. One notable effect of unlearning male-specific features is that the transformed images exhibit broader smiles. This is due to the sociological phenomenon wherein women tend to smile more than men in photographs~\citep{wondergem2012gender}. Furthermore, we detect an increase in eye make-up in images of females after unlearning. For more examples of these effects, see a larger collection of before/after unlearning pairs in Appendix~\ref{app:celeba-more}, where we also show an example of unlearning eyeglasses {and an example for nonsequential multiclass unlearning where we forget the composite class of male faces with dark hair (either black or brown). Interestingly, for the multiclass unlearning example, the composite class is forgotten via replacing the “male” attribute rather than the hair color attribute, suggesting that certain class-specific features are selectively forgotten while other shared features are retained (see Figure~\ref{fig:celeba-multi}).
} 

%Figure~\ref{fig:celeba} and 
We report in Table~\ref{tab:res} on the performance metrics for our methods, gradient surgery and feature unlearning for VAEs by \citet{moon2024feature} (for pseudocode see Algorithm~\ref{algo:vae_fu} in Appendix~\ref{app:algo}). We remark that gradient ascent (A) and ascent-descent (A-D) lead to numerical instability during unlearning generating white images. Table~\ref{tab:res} shows that UNO-S again achieves the fastest time to unlearn, followed by UNO. Even after spending $\sim20$ times more execution time than the time to unlearn with UNO, gradient surgery is unable to achieve the desired $\le2\%$ male faces in the generated images. After the $659$ allotted training steps gradient surgery is only able to reach $\sim4\%$ male faces. Only UNO and UNO-S have zero 5-step f-ratio implying they don't require more than $5$ steps on average to unlearn to the desired threshold. 
All algorithms result in similar values of FID, and the quality of the generated images is perceptually indistinguishable from the originally generated images, as seen in Figure~\ref{fig:celeba-transform}.

% \begin{figure}[!htp]
%     \centering
% \includegraphics[width=\linewidth]{plots/celeba.pdf}
%     \caption{Time to unlearn (left) and FID (right) for various unlearning algorithms applied on CelebA. The middle panel shows the fraction of generated images classified as "male", averaged over $10$ runs, as a function of training steps. Standard deviations over $10$ independent runs are shown as one-sided error bars in the left and right panels. For time to unlearn and FID, the rank of each algorithm is also indicated below its name, with $1$ being the best.}
%     \label{fig:celeba}
% \end{figure}

\subsection{ImageNet-1K}
We use a $675$M-parameter diffusion transformer DiT-XL/2~\citep{peebles2023scalable} that operates on a $4\times32\times32$-dimensional latent space, trained for $7$M steps on $1.28$M images at $256\times256$ resolution. We consider replacement unlearning and apply the conditional variants of UNO and UNO-S as described in Section~\ref{ssec:replace}. In particular, we attempt to unlearn images of golden retrievers (class $207$) by mapping them to images of labrador retrievers (class $208$), i.e. $c_t={208}$ and $c_f={207}$. This reduces the training to only two classes, rather than the full dataset. We employ $100$ training steps with a mini-batch size of $10$, and a learning rate of $\eta=10^{-4}$. To compute the time to unlearn, at each training step we generate samples only for the class $207$ with a (classifier-free) guidance scale of $8$ and determine what fraction of the samples are classified as golden retrievers. 

\begin{figure}[h]
    \centering
\includegraphics[width=\linewidth]{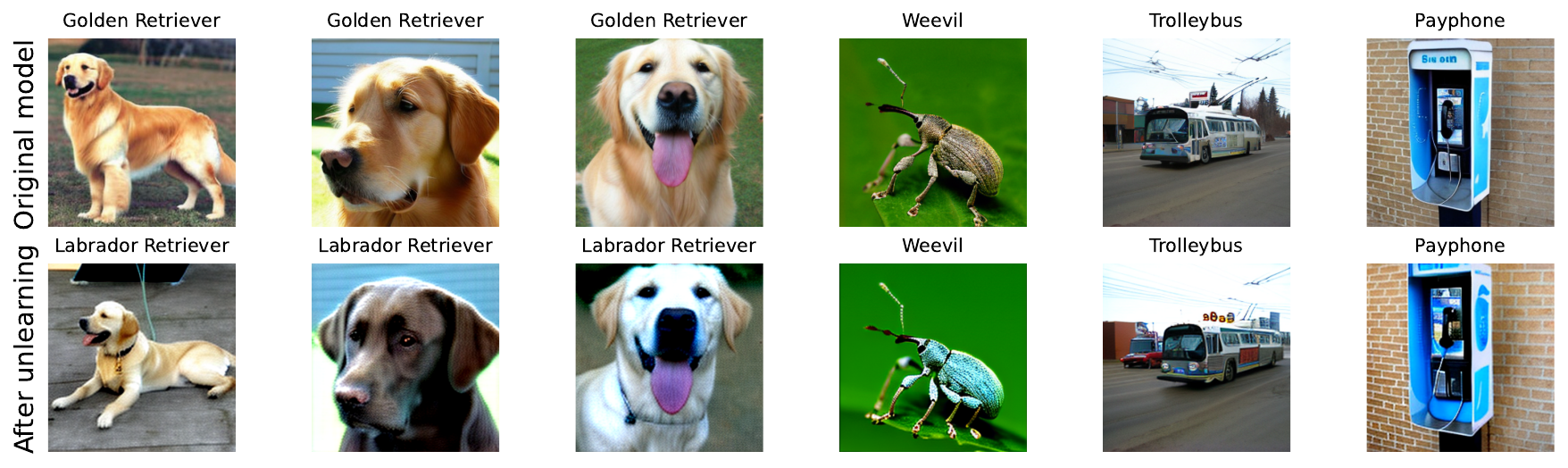}
    \caption{ImageNet-1K samples generated by the original model (top) and after replacement unlearning with UNO-S (bottom). We replace class $207$ (golden retriever) with class $208$ (labrador retriever), using identical noise inputs for the diffusion transformer. The labels classifying each image were provided by the pretrained classifier.}
    \label{fig:imagenet-transform}
\end{figure}

Figure~\ref{fig:imagenet-transform} shows that noise in the latent space that generated golden retrievers in the original model, generates labrador retrievers after unlearning with conditional UNO-S. Images belonging to other classes are significantly more changed than for CelebA, but remain in their respective classes. A larger collection of before/after unlearning pairs are provided in Appendix~\ref{app:imagenet-more}.       
Table~\ref{tab:res} summarizes the performance metrics for our methods, gradient surgery and concept erasure for diffusion models by \citet{gandikota2023erasing}. We provide pseudocode for all methods in Appendix~\ref{app:algo}. For concept erasure we use the two choices of erasure strength hyperparameter $\lambda=1$ and $\lambda=3$ labelled as ESD-u-1 and ESD-u-3 in \citet{gandikota2023erasing}. All algorithms produce images that are visually of the same quality and have comparable FID values. For ImageNet-1K the differences in performance are less pronounced. We believe that this is due to the simpler set-up of conditional unlearning with only two classes. We note that gradient ascent (A) and ascent–descent (A-D) suffer from catastrophic forgetting, and we therefore do not report their corresponding results. 
% SalUn exhibits similar behaviour, as reflected in its high FID score. 
We remark that an FID value of approximately $2.3$ was reported in \citet{peebles2023scalable}, whereas our FID values are around $12$. This is due to their significantly larger sample size of $50,000$ compared to our $22,000$ samples. Our algorithms outperform gradient surgery on time to unlearn, however, our algorithms are slower than the concept erasure algorithm ESD-u-3. Despite taking longer to unlearn, UNO and UNO-S require the least number of steps before unlearning, with the smallest 5-step f-ratios. 

\begin{figure}[h]
    \centering
\includegraphics[width=\linewidth]{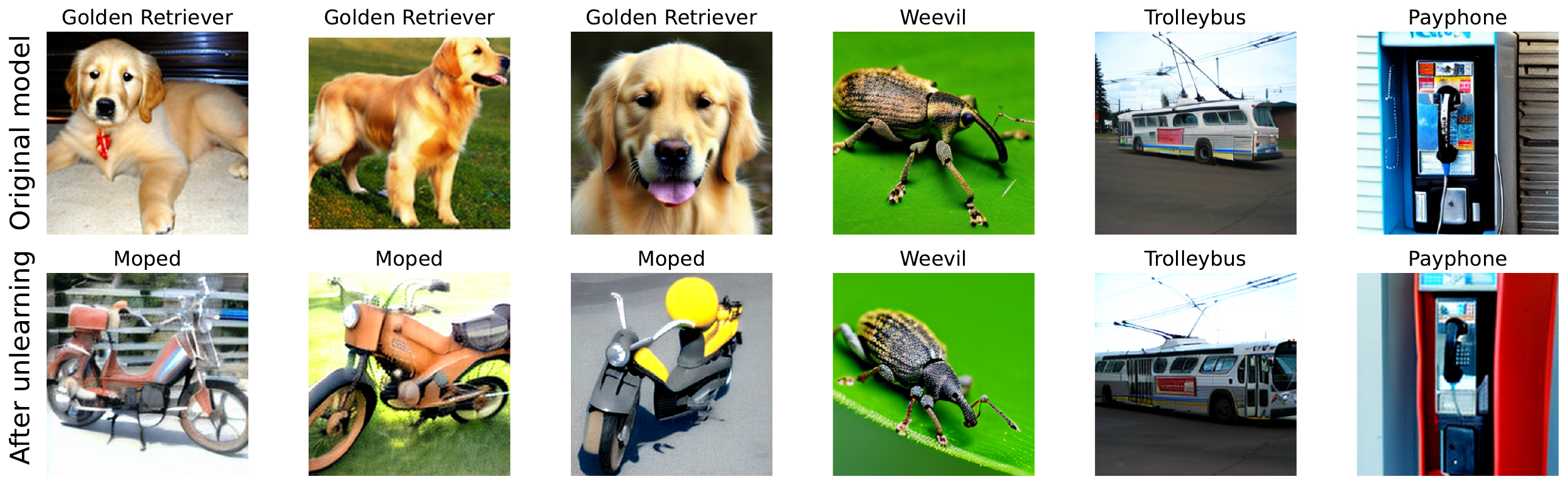}
 \caption{{ImageNet-1K samples generated by the original model (top) and after replacement unlearning  with UNO-S (bottom). We replace class $207$ (golden retriever) with class $665$ (moped), using identical noise inputs for the diffusion transformer. The labels classifying each image were provided by the pretrained classifier.}}
    \label{fig:imagenet-transform-moped}
\end{figure}

{So far, we have considered replacement unlearning where the forget and the retain classes share common features. We now show results where this is not the case. In particular, we attempt to unlearn images of golden retrievers (class $207$) by mapping them to images of mopeds (class $665$), i.e. $c_t={665}$ and $c_f={207}$. Figure~\ref{fig:imagenet-transform-moped} shows that UNO-S is capable of unlearning when the classes share very few features. Results for UNO (not shown) give equally good results.} 

\begin{figure}[!htp]
    \centering
    \includegraphics[width=0.65\linewidth]{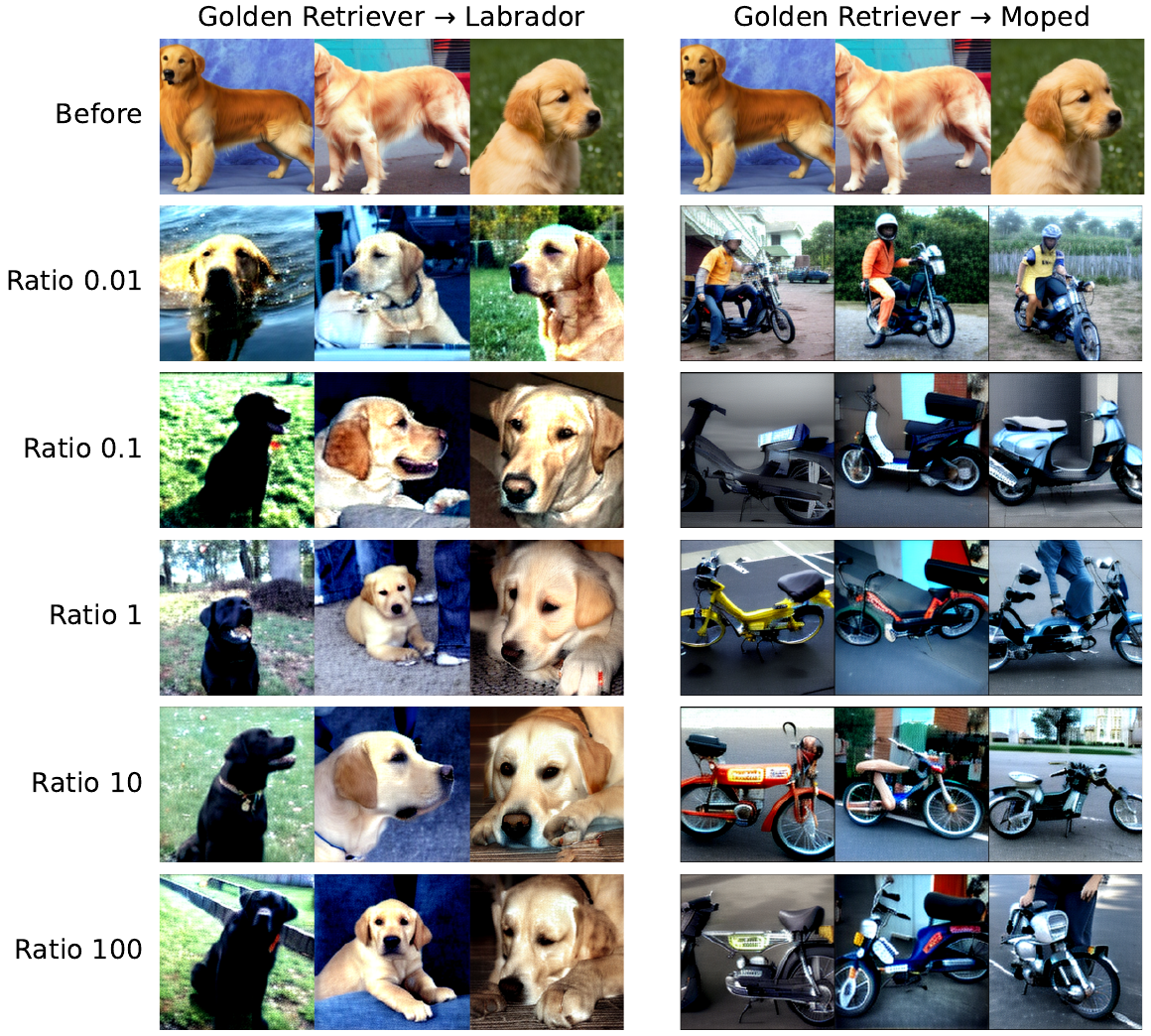}
    \caption{{ImageNet-1K samples generated by the original model (top) and after replacement unlearning with UNO-S for different ratios of the cardinalities of the retain and forget datasets. We show replacement unlearning of class $207$ (golden retriever) with class $208$ (labrador retriever) (left) and replacement unlearning of class $207$ (golden retriever) with class $665$ (moped) (right). We use identical noise samples for generating the before and after images.}}
    \label{fig:imagenet-ratios}
\end{figure}

{In the previous experiments unlearning was performed when we had the same number of training data for the retain and the forget class. It is natural to ask if our unlearning methods perform equally well when the cardinalities of the two sets vary significantly. In Figure~\ref{fig:imagenet-ratios} we show how UNO-S is able to perform successful replacement unlearning for ratios of the cardinalities of the retain dataset and the forget dataset, respectively. We consider extreme situations with the ratios varying from $0.01$ to $100$. The ratios of $0.01$ and of $100$ imply that one of the smaller of the two sets has only $13$ members out of the total number of $1,300$ available training samples per class. We show results for both, replacement unlearning where the classes share many features (golden retrievers to labrador retrievers) and where they do not share many common features (golden retrievers to mopeds). UNO and gradient surgery yield similar results (see Appendix~\ref{app:imagenet-more}). When only a few retain samples are available for replacement unlearning, e.g. for a ratio of $0.01$, the generated samples exhibit reduced variability (cf Figure~\ref{fig:imagenet-ratios-UNO} in Appendix~\ref{app:imagenet-more}).}

\begingroup
\begin{table}[htp]
\caption{Performance of various algorithms for class/feature unlearning with VAEs on MNIST and CelebA and diffusion transformers on ImageNet-1K. Each experiment is repeated $10$ times, and the standard deviations are shown in parentheses. FID values are calculated using $25,000$ samples for MNIST and CelebA and $22,000$ samples for ImageNet-1K. {\color{red}\ding{55}} indicates that the generated samples after unlearning are unrecognizably different from the original model. Bold indicates the best score among instances with acceptable FID. {\color{green!60!black}\ding{51}} indicates the generated samples after unlearning are perceptually indistinguishable from the original model in terms of visual fidelity. An asterisk (*) denotes cases where the algorithm failed to reach the target fraction of forget samples in the generated images within the allotted number of training steps. We only report on (A) and (A-D) for MNIST since they demonstrate catastrophic forgetting in all cases; for CelebA (A) and (A-D) result in NaN for FID, generating white images. The original models for MNIST, CelebA and ImageNet-1K have FID $20.7$, $166.3$ and $12.7$, respectively, and we are unlearning the digit $1$, faces identified as "male" and images of class $207$ (golden retriever), respectively.}

\label{tab:res}
\centering
\begin{tabular}{@{}ll@{\hspace{-1em}}c@{\hspace{-0.5em}}c@{\hspace{-0.2em}}c@{\hspace{-0.5em}}c@{}}
\toprule
\textbf{Dataset} & \textbf{Algorithm} & \hspace{-1em}Time to unlearn (s) $\downarrow$& FID $\downarrow$& $\Delta t$ (ms) & $5$-step f-ratio $\downarrow$\\
\midrule
\multirow{5}{*}{\shortstack{MNIST }}
  & Gradient ascent (A) & \textbf{0.024} (0.004) & 612.3 (4.9) {\color{red}\ding{55}} & 5.1 (0.47) & 0.00 (0.01)\\
  & Ascent descent (A-D) & 0.025 (0.006) & 266.9 (19.3) {\color{red}\ding{55}} & 5.1 (0.09) & 0.01 (0.01)\\
  & Gradient surgery (S) & 1.016 (0.907) & 23.0 (0.2) {\color{green!60!black}\ding{51}} & 7.1 (0.06) & 0.05 (0.03)\\
  & UNO & 0.055 (0.009) & \textbf{21.8} (0.2) {\color{green!60!black}\ding{51}} & 20.5 (0.27) & \textbf{0.00} (0.00)\\
  & UNO-S & 0.041 (0.010) & 21.8 (0.4) {\color{green!60!black}\ding{51}} & 15.1 (0.35) & \textbf{0.00} (0.00)\\
  & \citet{moon2024feature} & 0.065 (0.052) & 25.1 (0.4) {\color{green!60!black}\ding{51}} & 5.3 (0.04) & 0.03 (0.02)\\
\midrule
\multirow{3}{*}{\shortstack{CelebA}}
  & Gradient surgery (S) & 10.750$^*$ (3.362) & 176.0 (3.7) {\color{green!60!black}\ding{51}} & 18.3 (0.18) & 0.11 (0.05)\\
  & UNO & 0.527 (0.003) & 174.3 (1.6) {\color{green!60!black}\ding{51}} & 175.7 (0.91) & \textbf{0.00} (0.00)\\
  & UNO-S & \textbf{0.283} (0.040) & 177.1 (4.7) {\color{green!60!black}\ding{51}} & 101.0 (0.35) & \textbf{0.00} (0.00)\\
  & \citet{moon2024feature} & 7.827 (3.683) & \textbf{172.9} (0.8) {\color{green!60!black}\ding{51}} & 15.4 (0.04) & 0.14 (0.03)\\
\midrule
\multirow{3}{*}{\shortstack{ImageNet-1K}}
  & Gradient surgery (S) & 7.622 (2.211) & 12.3 (0.8) {\color{green!60!black}\ding{51}} & 712.4 (8.99) & 0.70 (0.28)\\
  & UNO & 6.361 (0.909) & 11.8 (0.8) {\color{green!60!black}\ding{51}} & 1817.3 (4.59) & \textbf{0.00} (0.00)\\
  & UNO-S & 6.495 (1.928) & 12.0 (0.8) {\color{green!60!black}\ding{51}} & 1273.4 (8.04) & 0.07 (0.12)\\
  & ESD-u-1~\citep{gandikota2023erasing} & 6.600 (1.299) & \textbf{11.6} (0.6) {\color{green!60!black}\ding{51}} & 362.6 (4.96) & 0.63 (0.20)\\
  & ESD-u-3~\citep{gandikota2023erasing} & \textbf{2.023} (0.677) & 13.1 (1.0) {\color{green!60!black}\ding{51}} & 348.8 (5.64) & 0.14 (0.10)\\
  % & SalUn~\citep{fan2024salun} & 4.411 (0.766) & 242.2 (3.2) {\color{red}\ding{55}} & 711.4 (3.36) & 0.13 (0.09)\\
\bottomrule
\end{tabular}

\end{table}

\begingroup
\setlength{\tabcolsep}{4pt} 

%%%%%%%%%%%%%%%%%%%%%%%%%%%%%%%%%%%%%%%%%%%%%%%%%%%%%%%%%%%%%%%%%%%%%%%
{
\subsection{Attacks on unlearned models}
\label{sec:attacks}

So far we have assessed the performance of our unlearning methods by assessing the quality of the generated data and if the model suppresses generation of samples classified as belonging to the forget class. It is well known that such behavioural suppression does not imply that the unlearned model has truly forgotten the data and that the forgotten data are not still encoded in the model parameters~\citep{hu2025unlearning,hatami2026class}. To investigate how far our unlearned models still encode the forget data we now look at loss-based membership attacks (MIA) and at relearning attacks. 
%Machine unlearning may introduce privacy vulnerabilities~\citep{chen2021machine}. In this section, we use some simple attacks to investigate whether the unlearning algorithms considered in this work exacerbate privacy risks.

\subsubsection{Loss-based membership inference attack}
\label{ssec:mia}

We run a loss-based membership inference attack (MIA), where lower loss implies membership, on the original and unlearned models. We run two versions of this attack: 1) the retain and forget classes are attacked separately where we use the training samples as members and validation samples as non-members, and 2) where we use the retain class as the members and forget class as non-members.  We report three key metrics. \textbf{ROC-AUC} can be interpreted as the probability that, for a randomly selected member–non-member pair, the member incurs a lower loss than the non-member. \textbf{Attack advantage} is the maximum difference between the true-positive and false-positive rates over all attack thresholds, and measures the best separation achievable between members and non-members. Attack advantage is related to the concept of balanced accuracy = $\frac{1 + \text{attack adv.}}{2}$, which is the probability an attacker with optimal thresholding can classify membership correctly. Finally, \textbf{TPR@1\% FPR} is the fraction of members correctly identified when the false-positive rate is constrained to 1\%. We also report 95\% percentile bootstrap confidence intervals for these quantities.

% \subsubsection*{Class-separated membership inference attack}
Table~\ref{tab:mia-class-sep} reports the results for class-separated membership inference attack where the classes are defined by the retain and forget data. We consider the training and the validation data as members and non-members, respectively. For both retain and forget classes, across all models, the ROC-AUC is close to $0.5$ implying minimal leakage under loss-based MIA. For MNIST only the original model shows modest leakage on the retain set with an attack advantage of $5.36$\% which is lowered by the unlearning algorithms, except for Moon et al. which remains close to the original model. 
For ImageNet-1K, we detect no change in the attack diagnostics across the different models, again showing that the unlearned models have the same privacy risk than the original model. However, note that for our ImageNet-1K experiments we only have $50$ validation samples available per class as we only considered labrador retrievers and golden retrievers. 
%For ImageNet-1K, the point estimates suggest slightly stronger leakage, with attack advantages of $12$–$16$\%. However, the confidence intervals are wide, reflecting the small sample size of only $50$ validation samples available per class. For the forget class all unlearned models stay close to the attack advantage of the original model. 
%
Overall, loss-based MIA reveals little evidence that unlearning increases membership leakage.
\begin{table*}[t]
  \centering
  \scriptsize
  \setlength{\tabcolsep}{3pt}
  \caption{Class-separated loss-based membership inference, for which lower loss predicts membership.
  Each entry reports the point estimate followed by a stratified 95\% percentile
  bootstrap confidence interval in brackets. Members and non-members were
  resampled independently at their original group sizes for 1,000 image-level
  bootstrap replicates. Attack advantage is
  $\max_{\tau}(\mathrm{TPR}_{\tau}-\mathrm{FPR}_{\tau})$. TPR@1\% FPR is the maximum
TPR attained at an empirical FPR no greater than 1\%.}
  \label{tab:mia-class-sep}
  \resizebox{\textwidth}{!}{%
  \begin{tabular}{llcccccc}
    \toprule
    & & \multicolumn{3}{c}{Retain} & \multicolumn{3}{c}{Forget} \\
    \cmidrule(lr){3-5}\cmidrule(lr){6-8}
    Dataset & Model
      & ROC-AUC [95\% CI]
      &\makecell{Attack Adv.(\%)\\ {[95\% CI]}}
      & \makecell{TPR@1\% FPR (\%)\\{[95\% CI]}}
      & ROC-AUC [95\% CI]
      &\makecell{Attack Adv.(\%)\\ {[95\% CI]}}
      & \makecell{TPR@1\% FPR (\%)\\{[95\% CI]}} \\
    \midrule
     & Original & 0.534 [0.526, 0.543] & 5.36 [4.52, 6.83] & 1.06 [0.80, 1.35] & 0.484 [0.459, 0.506] & 0.70 [0.53, 4.23] & 1.32 [0.53, 2.11] \\
   & Gradient ascent & 0.498 [0.489, 0.506] & 1.68 [0.59, 3.12] & 0.96 [0.69, 1.41] & 0.507 [0.483, 0.531] & 2.56 [1.23, 7.22] & 0.62 [0.18, 1.68] \\
   & Ascent descent & 0.493 [0.485, 0.502] & 1.15 [0.38, 2.37] & 1.13 [0.81, 1.48] & 0.490 [0.467, 0.515] & 0.53 [0.44, 5.46] & 1.15 [0.26, 2.38] \\
    MNIST & Moon et al. & 0.530 [0.522, 0.539] & 4.99 [4.35, 6.44] & 0.91 [0.62, 1.18] & 0.492 [0.468, 0.515] & 1.23 [0.70, 5.29] & 1.41 [0.44, 2.82] \\
    & UNO & 0.504 [0.496, 0.513] & 1.40 [0.96, 2.70] & 1.14 [0.81, 1.56] & 0.480 [0.457, 0.505] & 0.88 [0.44, 4.41] & 1.67 [0.53, 2.64] \\
     & UNO-S & 0.502 [0.493, 0.511] & 0.99 [0.63, 2.37] & 1.20 [0.83, 1.59] & 0.479 [0.455, 0.503] & 0.88 [0.44, 3.88] & 1.59 [0.70, 2.64] \\
   & Surgery & 0.506 [0.497, 0.514] & 1.61 [0.85, 3.07] & 1.17 [0.90, 1.60] & 0.467 [0.444, 0.492] & 0.53 [0.18, 3.70] & 0.79 [0.18, 1.67] \\
    \midrule
     & Original & 0.497 [0.490, 0.504] & 0.33 [0.12, 1.66] & 0.95 [0.68, 1.12] & 0.502 [0.493, 0.511] & 1.44 [0.63, 3.00] & 1.27 [0.90, 1.55] \\
     & Moon et al. & 0.495 [0.488, 0.502] & 0.32 [0.13, 1.41] & 0.90 [0.67, 1.10] & 0.509 [0.500, 0.518] & 1.94 [1.16, 3.65] & 1.08 [0.77, 1.38] \\
    CelebA & UNO & 0.497 [0.490, 0.505] & 0.32 [0.16, 1.59] & 0.90 [0.67, 1.16] & 0.503 [0.494, 0.511] & 1.73 [0.69, 3.11] & 1.30 [0.96, 1.63] \\
     & UNO-S & 0.497 [0.490, 0.504] & 0.35 [0.15, 1.57] & 0.89 [0.65, 1.16] & 0.501 [0.493, 0.510] & 0.76 [0.40, 2.46] & 1.15 [0.85, 1.56] \\
     & Surgery & 0.497 [0.489, 0.504] & 0.49 [0.19, 1.65] & 0.87 [0.66, 1.10] & 0.506 [0.498, 0.515] & 1.61 [0.87, 3.13] & 1.09 [0.80, 1.50] \\
    \midrule
     & Original & 0.556 [0.447, 0.672] & 16.00 [8.00, 36.00] & -- & 0.544 [0.424, 0.655] & 14.00 [8.00, 34.00] & -- \\
     & ESD-u-1 & 0.557 [0.451, 0.672] & 16.00 [8.00, 36.05] & -- & 0.543 [0.421, 0.660] & 14.00 [6.00, 34.00] & -- \\
     ImageNet-1K& ESD-u-3 & 0.556 [0.440, 0.664] & 16.00 [6.00, 36.00] & -- & 0.542 [0.421, 0.650] & 12.00 [6.00, 34.00] & -- \\
     & UNO & 0.557 [0.444, 0.671] & 16.00 [8.00, 38.00] & -- & 0.544 [0.431, 0.652] & 14.00 [8.00, 34.00] & -- \\
    & UNO-S & 0.557 [0.443, 0.668] & 16.00 [6.00, 38.00] & -- & 0.544 [0.433, 0.647] & 14.00 [8.00, 32.00] & -- \\
     & Surgery & 0.557 [0.448, 0.670] & 16.00 [6.00, 36.00] & -- & 0.544 [0.430, 0.658] & 14.00 [6.00, 36.00] & -- \\
    \bottomrule
  \end{tabular}%
  }
  \vspace{2pt}

  \begin{minipage}{0.98\textwidth}
    \footnotesize
    \textit{Notes.} The retain/forget targets are digits other than 1/digit 1
    for MNIST, Female/Male for CelebA, and ImageNet class 208/class 207,
    respectively. Available samples per side for retain/forget are 8,865/1,135 for
    MNIST, 11,409/8,458 for CelebA, and 50/50 for ImageNet. A dash indicates
    that 1\% FPR is not resolved for ImageNet-1K because its smallest nonzero FPR
    is $1/50=2\%$. The advantage-maximizing threshold is selected separately
    on each bootstrap ROC; its interval therefore describes empirical oracle
    advantage rather than performance at a calibrated threshold. ImageNet-1K
    intervals are much wider because of its small sample size.
  \end{minipage}
\end{table*}

Table~\ref{tab:dit-retain-forget-mia} reports the results for MIA when the retain and forget data are treated as members and non-members, respectively. We only report here on ImageNet-1K. In our MNIST and CelebA experiments we employed VAEs. However, VAEs exhibit an imbalanced loss across classes where the loss evaluated for the forget class is significantly different to the loss evaluated at the retain class, even for the original models. This violates the assumption that lower loss implies membership, necessary for the applicability of loss-based MIA. On the contrary, diffusion transformers (DiTs) have a more balanced loss distribution across retain and forget data. Like the class-separated membership inference attack, discussed above, we do not observe a significant difference in privacy risk across the unlearned models. The results indicate that the retain and forget classes are only weakly distinguishable using the loss-based membership score, with the ROC-AUC being close to $0.5$, the attack advantage being $\sim 6$\%, and the TPR being $1$--$4$\% at $1$\% FPR.  The attack advantage of approximately $6$\% implies that an optimal loss-based attacker can identify membership correctly $53$\% of the time, across all models, only marginally better than guessing randomly. 

\begin{table}[t]
  \centering
  \small
  \setlength{\tabcolsep}{8pt}
  \caption{Loss-based membership inference for the unlearned DiT models,
  treating retained class 208 training images as members and forgotten class
  207 training images as non-members. Attack advantage is
  $\max_{\tau}(\mathrm{TPR}_{\tau}-\mathrm{FPR}_{\tau})$. TPR@1\% FPR is the maximum
TPR attained at an empirical FPR no greater than 1\%. Brackets give 95\%
  bootstrap confidence intervals from $1000$ replicates.}
  \label{tab:dit-retain-forget-mia}
  \begin{tabular}{lccc}
    \toprule
    Model & ROC-AUC [95\% CI] & Attack adv. (\%) [95\% CI]
      & TPR@1\% FPR (\%) [95\% CI] \\
    \midrule
    ESD-u-1   & 0.528 [0.448, 0.557] & 6.00 [0.46, 10.62]
      & 1.08 [0.15, 2.31] \\
    ESD-3 & 0.530 [0.453, 0.559] & 6.31 [0.31, 10.92]
      & 2.46 [0.46, 4.62] \\
    UNO          & 0.526 [0.451, 0.554] & 5.85 [0.46, 10.15]
      & 1.08 [0.15, 3.08] \\
    UNO-S        & 0.528 [0.451, 0.557] & 5.69 [0.31, 10.77]
      & 3.54 [0.31, 5.69] \\
    Surgery      & 0.526 [0.454, 0.552] & 5.85 [0.31, 10.31]
      & 1.69 [0.31, 2.92] \\
    \bottomrule
  \end{tabular}
\end{table}

\subsubsection{Few-shot relearning attacks}
\label{ssec:relearn}

\begin{table}[t]
  \centering
  \small
  \caption{Normalized change in forgotten-class generation during a
  single-run of 20-shot relearning attack with $50$ update steps.}
  \label{tab:relearning-20shot}
  \begin{tabular}{llc}
    \toprule
    Dataset & Model
      & $\Delta p_F^{(50)}$ \\
    \midrule
     & Gradient ascent       & 0.000 \\
     & Ascent--descent       & 0.315 \\
     & Moon et al.           & 0.261 \\
    MNIST & UNO                   & 0.270 \\
    & UNO-S                 & 0.288 \\
   & Surgery               & 0.288 \\
  & Retraining oracle     & 1.000 \\
    \midrule
     & Moon et al.          & 1.443 \\
   & UNO                  & 1.357 \\
    CelebA & UNO-S                & 1.529 \\
   & Surgery              & 1.329 \\
     & Retraining oracle    & 1.000 \\
    \bottomrule
  \end{tabular}
  \vspace{2pt}

  \begin{minipage}{0.96\linewidth}
    \footnotesize
    \textit{Notes.} Each entry is computed from one attack run using
    128 generated samples per diagnostic. Values can exceed
    one when an attacked model's change in $p_F$ is larger than the oracle's
    change over the same $50$ updates.
  \end{minipage}
\end{table}

A different diagnostic tool to evaluate whether an unlearned model has unlearned the forget data is given by relearning attacks. In relearning attacks we take an unlearned model and retrain it on a small amount of the forget data. We compare the (normalized) change in the f-ratios (fraction of forget data in a fixed number of generated samples) of the unlearned model and a retraining oracle (i.e. a model that was only trained on retain data) after relearning on $20$ forget samples for $50$ steps. In particular, we define
\begin{align}
\Delta p_F^{(50)} = \frac{|p_{F}^{(50)} - p_{F}^{(0)}|}{|p_{F,r}^{(50)} - p_{F,r}^{(0)}|} \,,
\label{eq:delta-p}
\end{align}
where $p_F^{(t)}$ and $p_{F,r}^{(t)}$ are the fractions of generated samples classified as the forgotten class after $t$ attack updates for an unlearned model and for a retraining oracle, respectively. Note that, for an ideal classifier $p_{F}^{(0)}=p_{F, r}^{(0)}=0$, but we keep them in \eqref{eq:delta-p} in order to account for classifier errors. If the difference in the f-ratios obtained from the unlearned model is significantly larger than that of the control, it shows that the unlearned model still contains information about the forget dataset and did not effectively unlearn. This helps us to distinguish between behavioral suppression and "true unlearning" where the influence of the forget data on the model parameters has been removed. We report results for relearning attacks in Table~\ref{tab:relearning-20shot}. It is seen that relearning behavior differs sharply across datasets. For MNIST, all unlearned models produce less forgotten data in generation than the retraining oracle, indicating persistent resistance to relearning. For CelebA, every unlearned model exceeds the oracle’s recovery. These results show that resistance to relearning is strongly dataset-dependent and cannot be inferred solely from a model’s immediate post-unlearning behavior. For VAEs trained on CelebA, the forgotten attribute remains latent and can be restored more readily than with a retraining oracle. We do not report results for ImageNet-1K due to the prohibitive computational cost of training a diffusion transformer from scratch. However, we note that, in our experiments, concept erasure~\citep{gandikota2023erasing} was particularly resistant to few-shot relearning attacks.
}
%%%%%%%%%%%%%%%%%%%%%%%%%%%%%%%%%%%%%%%%%%%%%%%%%%%%%%%%%%%%%%%%%%%%%%

{
\subsection{Influence of the orthogonalization on unlearning}
%Cosine similarity between loss gradients during unlearning}
Our unlearning methods involve an additional term in the loss function \ref{eq:loss-uno} promoting the orthogonality between the gradients $\gr$, evaluated at the retain data, and $\gf$, evaluated at the forget data. We show here how during unlearning indeed the two gradients $\gr$ and $\gf$ become more and more misaligned during training. We investigate the degree of misalignment during unlearning for UNO and UNO-S as well as for gradient surgery as a control method that does not involve any additional term in the loss function. Figure~\ref{fig:cosine} illustrates the evolution of an unlearning run for gradient surgery and for UNO and UNO-S on the ImageNet-1K replacement task where we replace class $207$ (golden retriever) with class $208$ (labrador retriever) using all available $1,300$ training samples for both classes. We show the f-ratio versus the cosine similarity between the two loss gradients, as defined in \eqref{eq:cossim}, during unlearning until a threshold f-ratio of $\tau=0.18$ is reached. We show the averaged cosine similarity over $10$ independent instances of unlearning. At each realization the cosine similarity is computed using gradients evaluated over the full retain and forget validation datasets to avoid unwarranted fluctuations. Prior to unlearning, the f-ratio is $1$ and the cosine similarity is measured to be $\cos\vartheta\approx 0.8$. During unlearning the f-ratio decreases while (on average) the cosine similarity decreases as well for all three methods. The results for gradient surgery which does not explicitly promote orthogonality of the retain and forget gradients suggests that unlearning and the orthogonality of the retain and forget gradients are concurrent. The inclusion of the orthogonality promoting term in the loss function used for UNO and UNO-S then implies, as Figure~\ref{fig:cosine} shows, a significant speed up in unlearning. 

Prior to unlearning the two terms in the loss function~\eqref{eq:loss-uno} are of equal magnitude. After only 3 update steps the orthogonality term is two orders of magnitude smaller than the retain data loss term. Any further gradient step hence aims at decreasing the retain data loss term. The cosine similarity does not decrease any further due to the small size of $\beta_o=0.02$. This suggests that to achieve efficient unlearning the two gradients do not need to be optimally misaligned with $\gr\cdot \gf=0$. The achieved misalignment after the first three steps in UNO and UNO-S allows for a much faster "learning" of the retain data. Recall that a retraining oracle would perform standard gradient descent on the loss function~\eqref{eq:loss-uno} with $\beta_o=0$, but would take significantly longer than gradient surgery. It appears that the additional information about the local geometry and curvature of the loss landscape via the Hessian in the UNO and the UNO-S updates, compared to standard gradient descent of the retraining oracle which only encodes information about the gradient, enables this speed up. % by guiding the decsent . 

%The exact mechanism of how and why the weak orthogonality constraint in the loss function~\eqref{eq:loss-uno} achieves an advantageous preconditioning of the model weights in the first few steps allowing for fast gradient descent along the retain data loss term remains an open question. 

For an analytical treatment of the effect of orthogonalization in a linear model see Appendix~\ref{app:lin}.

\begin{figure}[!htp]
    \centering
\includegraphics[width=0.6\linewidth]{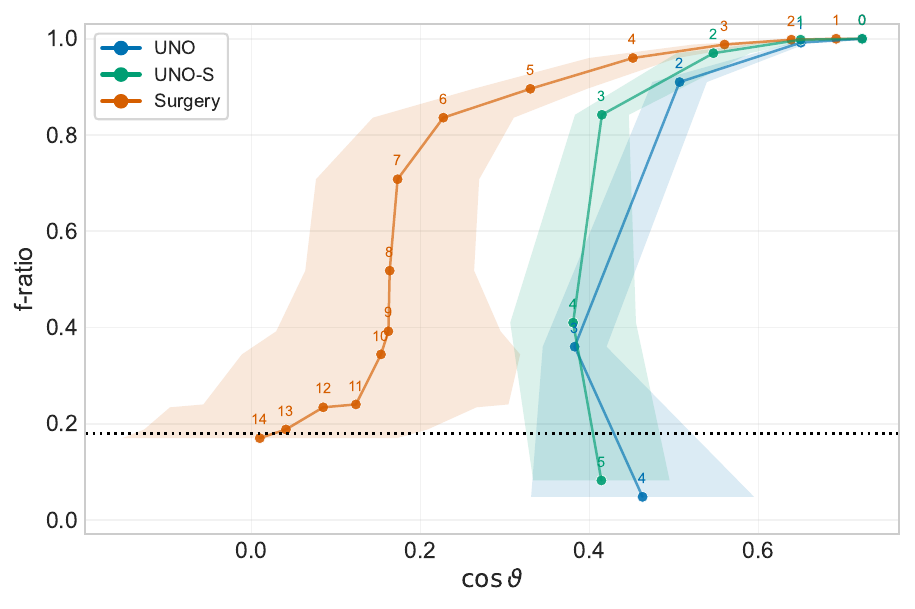}
    \caption{{The f-ratio versus the cosine similarity between loss gradients for various algorithms until a threshold f-ratio of $\tau=0.18$ is reached. The data here correspond to replacement unlearning of golden retrievers with labrador retrievers, discussed in Section~\ref{sec:res}. We show the average cosine similarity and f-ratio for $10$ independent realizations, together with confidence intervals indicating variability in the cosine similarity. The numbers on each data point refer to the respective update steps.}}
    \label{fig:cosine}
\end{figure}

%%%%%%%%%%%%%%%%%%%%%%%%%%%%%%%%%%%%%%%%%%%%%%%%%%%%%%%%%%%%%%%%%%%%%%%%%%
{
\subsection{Scalability}
\label{sec:scalability}
The additional regularization term of the UNO loss function~\eqref{eq:UNO} implies the evaluation of second-order derivatives in the update step. This additional computational overload renders a single update step of loss-gradient orthogonalization based methods, UNO and UNO-S, to have a significantly higher cost compared to other methods, as shown in Table~\ref{tab:res}. 
%shows that loss-gradient orthogonalization incurs substantially higher computational costs per update step compared to other methods. 
However, its overall unlearning time may be lower. In order to understand the computational cost of orthogonalization it suffices to look at the quantity
% $J=\nabla_\theta(\gr\cdot\gf)$, which is equivalent to 
${\rm{H}}_r\bgf+{\rm{H}}_f\bgr$ in the UNO update step \eqref{eq:hess-update}, where we recall that ${\rm{H}}_r, {\rm{H}}_f$ are the Hessians evaluated at the retain and forget data, respectively, as defined in~\eqref{eq:Hess_rf}. In order to compute the parameter update, however, we do not need to compute the full Hessians but rather only the Hessian-vector products (HVP). Remarkably, for a model that has cost $C$ associated with a single forward pass, HVP scales only as $O(C)$ with automatic differentiation (see~\citep{pearlmutter1994fast}). This renders orthogonalization scalable in practice. In particular, we never store or compute the full Hessians. In Figure~\ref{fig:complexity} we show the memory and time usage of various algorithms on an A100 GPU against the forward cost of the models. We record memory usage (left) as the difference between peak GPU memory used during an update step and GPU memory occupied immediately before the step, as well as the time $\Delta t$ required for a single update (right). We omit UNO-S and ascent-descent in Figure~\ref{fig:complexity} due to their cost profiles being identical to UNO (in memory) and gradient ascent (in memory and time) respectively. The sublinear scaling of UNO for both memory usage and computational time suggests favorable scalability, although it likely reflects several complex interacting factors, including differences in model architecture and memory-use efficiency. Low-rank adaptations~\citep{hu2021lora} of these unlearning algorithms could substantially reduce memory and computational costs, but we do not pursue them here.}

\begin{figure}[!htp]
\centering\includegraphics[width=\linewidth]{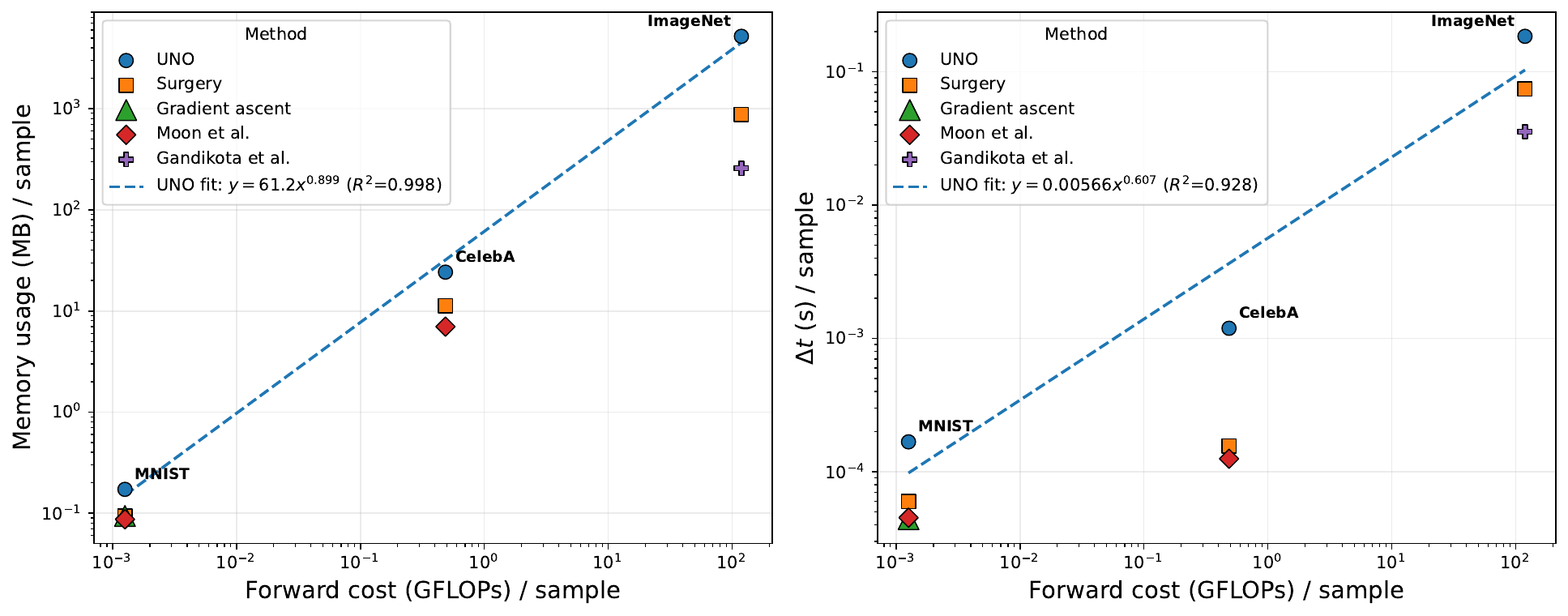}
    \caption{Computational complexity of unlearning methods across model scales as a function of the cost of a single model forward pass for a single sample, measured in GFLOPs. Left: memory usage per sample, defined as the difference between the peak GPU memory occupied during an update step and the memory occupied immediately before that step, divided by the training batch size. Right: $\Delta t$ per sample, defined as the duration of a single update step divided by the training batch size. Both panels use logarithmic axes. The dashed lines show power-law fits to the UNO results.}
    \label{fig:complexity}
\end{figure}
}

%%%%%%%%%%%%%%%%%%%%%%%%%%%%%%%%%%%%%%%%%%%%%%%%%%%%%%%%%%%%%%%%%%%%%%

%%%%%%%%%%%%%%%%%%%%%%%%%%%%%%%%%%%%%%%%%%%%%%%%%%%%%%%%%%%%%%%%%%%%%%
{
\section{Discussion}
\label{sec:disc}
We advance the gradient surgery paradigm for machine unlearning by introducing two new algorithms UNO and UNO-S and their conditional variants. UNO and UNO-S are constructed by augmenting the usual loss function by an additional term promoting the orthogonality between the gradients evaluated at the retain and forget data, respectively. This inclusion, we have shown, significantly increases the speed of unlearning by encoding information about the local geometry of the loss landscape. This strategy is designed to be model-agnostic and applicable across architectures. We show that UNO and UNO-S do not suffer from catastrophic forgetting. 
%, and are substantially faster than gradient surgery for unconditional generative models. 
UNO-S outperforms all other algorithms for unconditional generative models, and can be $1.3$ - $1.9$ times faster than UNO at unlearning. For conditional unlearning, UNO and UNO-S achieve comparable performance, and among methods that avoid catastrophic forgetting, only aggressive concept erasure (ESD-u-3) is faster. In all cases, UNO and UNO-S require the fewest parameter update steps to unlearn. We have demonstrated the efficiency of our algorithms across datasets and generative models of increasing complexity. We have also shown that incorporating information from a classifier that distinguishes between desirable and undesirable data can accelerate gradient surgery; see Appendix~\ref{app:un-w-cls}. 
% Furthermore, our experiments indicate that UNO and UNO-S are robust to hyperparameter selection. In particular, we used the same hyperparameter values for both MNIST and CelebA; see Table~\ref{tab:param}.

We have shown that UNO and UNO-S can be used for unlearning classes in situations where the retain and the forget datasets share many common features and in situations when they do not. Moreover, unlearning was achieved for a wide range of ratios between the cardinalities of the two classes. We showed that UNO and UNO-S can be used for both single class and multiclass unlearning.

We have mainly addressed here unlearning in the sense of behavioural suppression by monitoring the generated data of the unlearned model. However, our results on relearning attacks show that the unlearned model may regain the lost knowledge quicker than a retain-only retraining oracle suggesting that the unlearned model still encodes information about the forget dataset. On the other hand, we showed that our unlearning algorithms do not increase privacy risks of the original model under a loss-based membership inference attacker. 

It is straightforward to conceptualize low-rank adapted~\citep{hu2021lora, xu2023qa} variants of the unlearning algorithms presented here. Such modifications are essential for enabling efficient unlearning in large-scale generative models, and we leave their exploration to future research. The CelebA experiments show that unlearning via UNO and UNO-S can easily produce male-to-female face filters. Applications for designing a broader range of filters is an interesting topic for further exploration. Machine learning models used to simulate or predict physical systems, such as climate models, often generate unphysical states~\citep{lai2024machine}. A related challenge arises in video-generation models such as Sora~\citep{kang2024far}, which may produce physically implausible outputs. An interesting direction for future work is to investigate whether unlearning methods can be used to suppress such unphysical generations.}

\section*{Acknowledgments} The authors would like to acknowledge support from the Australian Research Council under
Grant No. DP220100931. We wish to thank Yiming Ying and Mehrtash Harandi for valuable discussions. 

% \newpage
\bibliography{references}
\bibliographystyle{tmlr}  

%%%%%%%%%%%%%%%%%%%%%%%%%%%%%%%%%%%%%%%%%%%%%%%%%%%%%%%%%%%%
\newpage
\appendix
%%%%%%%%%%%%%%%%%%%%%%%%%%%%%%%%%%%%%%%%%%%%%%%%%%%%%%%%%%%%%%%%%%%%%%%%%%%%%%

\section{Experiment setup}\label{app:setup}
In this section we provide additional details of the experimental setup used to produce the reported
results. 
\subsection{VAE loss functions and training data}
We document here the loss functions used to train the original model. For each input image $x$, the encoder outputs $\mu(x) \in \mathbb{R}^{d_z}$ and $\sigma(x) \in \mathbb{R}^{d_z}$, which parameterize the approximate posterior distribution. Here $d_z$ is the latent dimension. The corresponding reconstruction of $x$ by the decoder is denoted by $\bar{x}$,
 with $\bar{x}_i$ referring to its $i$-th pixel. 

The VAE used as the original model for MNIST was trained using the loss function
\begin{equation}
\begin{aligned}
    \mathcal{L}_{\rm MNIST}=\frac{1}{|\mathcal D|}\sum_{x\in \mathcal D}&\Big[-\sum_{i=1}^{784} ( x_i \log(\bar{x}_i) + (1 - x_i) \log(1 - \bar{x}_i) )\\ &+ \frac{1}{2} \sum_{i=1}^{d_z} 
\left( \mu_i^2(x) + \sigma_i^2(x) - \log \sigma_i^2(x) - 1 \right)\Big].
\end{aligned}\label{eq:loss-mnist}
\end{equation}
The $60,000$ training images were normalized such that pixel values lie in $[0, 1]$, following standard practice.  

The VAE used as the original model for CelebA was trained using the loss function
\begin{equation}
\begin{aligned}
    \mathcal{L}_{\rm CelebA}=\frac{1}{|\mathcal D|}\sum_{x\in \mathcal D}\left[\|x-\bar{x}\|^2+ \frac{1}{2} \sum_{i=1}^{d_z} 
\left( \mu_i^2(x) + \sigma_i^2(x) - \log \sigma_i^2(x) - 1 \right)\right].
\end{aligned}\label{eq:loss-celeba}
\end{equation}
We worked with $202,599$ cropped and aligned images in CelebA which originally have resolution $178\times178$ pixels. We downsampled these images to $64\times64$ resolution for training.  
\subsection{Hyperparameters}
Table~\ref{tab:param} lists the hyperparameters used in the unlearning experiments presented here. Here $\eta$ is the learning rate, $K$ is the number of training steps executed, $\beta_o$ is the weight for the orthogonalization loss term in \eqref{eq:loss-uno}, $\beta_h$ is the weight for the KL divergence loss term in \eqref{eq:loss-h}, $\alpha$ is a small positive threshold for stable computation of the KL divergence in \eqref{eq:loss-h}, $B$ is the batch size, and $N_{\rm FID}$ is the number of samples used for calculating FID. We use $N_g=B$ for all the algorithms in Appendix~\ref{app:h}, which determines the number of samples to be generated using the generative model. Each method was tested $10$ times for each dataset. For MNIST, FID was computed using features extracted from the classifier model, whereas for CelebA, features were computed using the InceptionV3 model~\citep{szegedy2016rethinking}.   All experiments were done on an A100 GPU provided by Google Colab. 
 
\begingroup
\setlength{\tabcolsep}{4pt} 
\begin{table}[h]
\caption{Experiment hyperparameters}
\label{tab:param}
\centering
\begin{tabular}{@{}llccccccc@{}}
\toprule
\textbf{Dataset} & \textbf{Algorithm} & $\eta$ & $K$ & $B\beta_o$ & $B\beta_h$ & $\alpha$ & $B$ & $N_{\rm FID}$ \\
\midrule
\multirow{5}{*}{\shortstack{MNIST \\ (Class: 1)}}
  & Gradient ascent (A)     & $10^{-3}$ & 530 & -      & -      & -         & 128 & 25,000 \\
  & Ascent descent (A-D)    & $10^{-3}$ & 530 & -      & -      & -         & 128 & 25,000 \\
  & Gradient surgery (S)    & $10^{-3}$ & 530 & -      & -      & -         & 128 & 25,000 \\
  & UNO                     & $10^{-3}$ & 530 & $10^3$ & -      & -         & 128 & 25,000 \\
  & UNO-S                   & $10^{-3}$ & 530 & $10^3$ & -      & -         & 128 & 25,000 \\
  & H                       & $10^{-3}$ & 530 & -      & $10^3$ & $10^{-8}$         & 128 & 25,000 \\
  & \shat                   & $10^{-3}$ & 530 & -      & $10^3$ & $10^{-8}$ & 128 & 25,000 \\
  & \unoh                   & $10^{-3}$ & 530 & $10^3$ & $10^3$ & $10^{-8}$         & 128 & 25,000 \\
  & \unosh                  & $10^{-3}$ & 530 & $10^3$ & $10^3$ & $10^{-8}$ & 128 & 25,000 \\
  & \citet{moon2024feature} & $10^{-3}$ & 530 & - & - & - & 256 & 25,000\\
\midrule
\multirow{3}{*}{\shortstack{CelebA\\ (Feature: Male)}}
  & Gradient surgery (S)    & $10^{-3}$ & 659 & -      & -      & -         & 128 & 25,000 \\
  & UNO                     & $10^{-3}$ & 659 & $10^3$ & -      & -         & 128 & 25,000 \\
  & UNO-S                   & $10^{-3}$ & 659 & $10^3$ & -      & -         & 128 & 25,000 \\
  & H                       & $10^{-3}$ & 659 & -      & $10^3$ & $10^{-8}$         & 128 & 25,000 \\
  & \shat                   & $10^{-3}$ & 659 & -      & $10^3$ & $10^{-8}$ & 128 & 25,000 \\
  & \unoh                   & $10^{-3}$ & 659 & $10^3$ & $10^3$ & $10^{-8}$         & 128 & 25,000 \\
  & \unosh                  & $10^{-3}$ & 659 & $10^3$ & $10^3$ & $10^{-8}$ & 128 & 25,000 \\
  & \citet{moon2024feature} & $10^{-3}$ & 659 & - & - & - & 256 & 25,000\\
\midrule
\multirow{3}{*}{\shortstack{ImageNet-1K\\ (Class: $207$)}}
 &Gradient surgery (S)    & $10^{-4}$ & 100 & -      & -      & -         & 10 & 22,000 \\
  & UNO                     & $10^{-4}$ & 100 & $2\times10^{-1}$ & -      & -         & 10 & 22,000 \\
  & UNO-S                   & $10^{-4}$ & 100 & $2\times10^{-1}$ & -      & -         & 10 & 22,000 \\
  & \citet{gandikota2023erasing}                  & $10^{-4}$ & 100 & - & -      & -         & 10 & 22,000 \\
  % & SalUn~\citep{fan2024salun}                  & $10^{-4}$ & 100 & - & -      & -         & 10 & 22,000 \\
\bottomrule
\end{tabular}
\vspace{0.5em}
\end{table}
\endgroup
Note that, we have used twice as large batch-sizes for ~\citet{moon2024feature} compared to the other methods to ensure the same amount of data gets processed during training, since, for the other methods (except for gradient ascent) minibatches of equal size ($B$) from both of the retain and forget datasets are processed during training.

Following \citet{moon2024feature}, we set $\alpha_{\rm FU}=3$ for Algorithm~\ref{algo:vae_fu}. For Algorithm~\ref{algo:gandikota}, we use $\lambda=1$ and $\lambda=3$, denoted by ESD-u-1 and ESD-u-3, respectively, in Table~\ref{tab:res}, following \citet{gandikota2023erasing}.
% Finally, following \citet{fan2024salun}, we set $\beta_{\rm retain}=10^{-3}$ and $\rho=0.5$ for Algorithm~\ref{algo:salun}. 
We use the Adam optimizer~\citep{kingma2015adam} for MNIST and CelebA, and AdamW optimizer~\citep{loshchilov2017decoupled} for ImageNet-1K.

\subsection{Model sizes}
The VAE models for MNIST and CelebA have $632{,}788$ and $8{,}742{,}659$ parameters with latent dimension $d_z=2$ and $d_z=512$, respectively. The classifier models for MNIST and CelebA have $159{,}410$ and $2{,}190{,}913$ parameters, respectively. For the exact model implementations, please refer to the code linked in Section~\ref{sec:intro}. The VAEs were trained for $200$ epochs on $60,000$ and $202,599$ images in MNIST and CelebA, respectively. The classifiers were trained for $10$ epochs on these datasets. For ImageNet-1K, we use the DiT-XL/2 diffusion transformer with $675.13$M parameters, trained for $7$M steps on $256\times256$ images; see \citep{peebles2023scalable} for details. For the ImageNet-1K classifier, we use microsoft/swinv2-tiny-patch4-window8-256~\citep{liu2022swin}, which has $28$M parameters.

%%%%%%%%%%%%%%%%%%%%%%%%%%%%%%%%%%%%%%%%%%%%%%%%%%%%%%%%%%%%%%%%%%%%%%%%%%%%%%
\newpage
\section{Pseudocode for unlearning algorithms}\label{app:algo}
This section presents the pseudocode for the unlearning algorithms used in this work.

\subsection{Gradient ascent} Algorithms~\ref{algo:A} and~\ref{algo:A-D} describe the gradient ascent (A), and alternating gradient ascent-descent (A-D), respectively.

\subsection{Gradient surgery}
Algorithms~\ref{algo:SA} and \ref{algo:S} describe gradient surgery with ascent in the forget direction (SA) and descent in the retain direction (S), respectively; in particular, the former appears in~\citet{bae2023gradient}.

\subsection{UNO and UNO-S}
Algorithms~\ref{algo:UNO} and \ref{algo:UNO-S} describe unlearning via orthogonalization (UNO), and alternating orthogonalization and surgery (UNO-S), respectively.

\subsection{Classifier-assisted unlearning}\label{app:h}
Algorithms~\ref{algo:Sh}, \ref{algo:UNOh}, \ref{algo:UNO-Sh}, and \ref{algo:H} describe \shat, \unoh, \unosh, and histogram unlearning, respectively. While in practice it is sufficient for a binary classifier to output a logit or probability, for simplicity of presentation we assume the classifier outputs $1$ for retain samples and $0$ otherwise.

%%%%%%%%%%%%%%%%%%%%%%%%%%%
\begin{algorithm}[!htp]
\caption{Gradient ascent (A)}\label{algo:A}
\begin{algorithmic}[1]
\State \textbf{Input:} Loss function $\mathcal{L}$,  forget dataset $\mathcal{D}_f$, trained model requiring unlearning $\mathcal{M}_\theta$, learning rate $\eta$, number of training steps $K$, batch size $B$.
\State \textbf{Output:} Updated model parameters $\theta$.
\For{$k = 1$ to $K$}
    \State Acquire mini-batch $D_f$ of size $B$ from $\mathcal{D}_f$.
    \State $\gf \gets \frac{1}{B}\sum_{x\in D_f}\nabla_{\theta} \mathcal{L}(\mathcal M_{\theta}, x)$
    \State $\theta \gets \theta + \eta  \gf$
\EndFor
\State \Return $\theta$
\end{algorithmic}
\end{algorithm}
%%%%%%%%%%%%%%%%%%%%%%%%%%%

%%%%%%%%%%%%%%%%%%%%%%%%%%%
\begin{algorithm}[!htp]
\caption{Alternating gradient ascent and descent (A-D)}\label{algo:A-D}
\begin{algorithmic}[1]
\State \textbf{Input:} Loss function $\mathcal{L}$, retain dataset $\mathcal{D}_r$, forget dataset $\mathcal{D}_f$, trained model requiring unlearning $\mathcal{M}_\theta$, learning rate $\eta$, number of training steps $K$, batch size $B$.
\State \textbf{Output:} Updated model parameters $\theta$.
\For{$k = 1$ to $K$}
    \State Acquire retain and forget mini-batches $D_r, D_f$  of size $B$ from $\mathcal{D}_r, \mathcal{D}_f$ respectively.
    \If {$k$ is odd}
    \State $\gf \gets \frac{1}{B}\sum_{x\in D_f}\nabla_{\theta} \mathcal{L}(\mathcal M_{\theta}, x)$
    \State $\theta \gets \theta + \eta  \gf$
    \Else
    \State $\gr \gets \frac{1}{B}\sum_{x\in D_r}\nabla_{\theta} \mathcal{L}(\mathcal M_{\theta}, x)$
    \State $\theta \gets \theta - \eta  \gr$
    \EndIf
\EndFor
\State \Return $\theta$
\end{algorithmic}
\end{algorithm}
%%%%%%%%%%%%%%%%%%%%%%%%%%%

%%%%%%%%%%%%%%%%%%%%%%%%%%%%%%%%%%%%

%%%%%%%%%%%%%%%%%%%%%%%%%%%
\begin{algorithm}[!htp]
\caption{Gradient surgery with ascent in forget direction (SA)}\label{algo:SA}
\begin{algorithmic}[1]
\State \textbf{Input:} Loss function $\mathcal{L}$, retain dataset $\mathcal{D}_r$, forget dataset $\mathcal{D}_f$, trained model requiring unlearning $\mathcal{M}_\theta$, learning rate $\eta$, number of training steps $K$, batch size $B$.
\State \textbf{Output:} Updated model parameters $\theta$.
\For{$k = 1$ to $K$}
    \State Acquire retain and forget mini-batches $D_r, D_f$  of size $B$ from $\mathcal{D}_r, \mathcal{D}_f$ respectively.
    \State  $\gr \gets \frac{1}{B}\sum_{x\in D_r}\nabla_\theta \mathcal{L}(\mathcal M_\theta, x)$
    \State $\gf \gets \frac{1}{B}\sum_{x\in D_f}\nabla_{\theta} \mathcal{L}(\mathcal M_{\theta}, x)$
    \State $\gf \gets \gf - \frac{\gr \cdot \gf}{\|\gr\|^2}\gr$
    \State $\theta \gets \theta + \eta \gf$
\EndFor
\State \Return $\theta$
\end{algorithmic}
\end{algorithm}
%%%%%%%%%%%%%%%%%%%%%%%%%%%

%%%%%%%%%%%%%%%%%%%%%%%%%%%
\begin{algorithm}[!htp]
\caption{Gradient surgery with descent in retain direction (S)}\label{algo:S}
\begin{algorithmic}[1]
\State \textbf{Input:} Loss function $\mathcal{L}$, retain dataset $\mathcal{D}_r$, forget dataset $\mathcal{D}_f$, trained model requiring unlearning $\mathcal{M}_\theta$, learning rate $\eta$, number of training steps $K$, batch size $B$.
\State \textbf{Output:} Updated model parameters $\theta$.
\For{$k = 1$ to $K$}
    \State Acquire retain and forget mini-batches $D_r, D_f$  of size $B$ from $\mathcal{D}_r, \mathcal{D}_f$ respectively.
    \State  $\gr \gets \frac{1}{B}\sum_{x\in D_r}\nabla_\theta \mathcal{L}(\mathcal M_\theta, x)$
    \State $\gf \gets \frac{1}{B}\sum_{x\in D_f}\nabla_{\theta} \mathcal{L}(\mathcal M_{\theta}, x)$
    \State $\gr \gets \gr - \frac{\gr \cdot \gf}{\|\gf\|^2}\gf$
    \State $\theta \gets \theta - \eta  \gr$
\EndFor
\State \Return $\theta$
\end{algorithmic}
\end{algorithm}
%%%%%%%%%%%%%%%%%%%%%%%%%%%

%%%%%%%%%%%%%%%%%%%%%%%%%%%%%%%%%%%%%%%

%%%%%%%%%%%%%%%%%%%%%%%%%%%
\begin{algorithm}[!htp]
\caption{Unlearning via orthogonalization (UNO)}\label{algo:UNO}
\begin{algorithmic}[1]
\State \textbf{Input:} Loss function $\mathcal{L}$, retain dataset $\mathcal{D}_r$, forget dataset $\mathcal{D}_f$, trained model requiring unlearning $\mathcal{M}_\theta$, weight for orthogonalization loss term $\beta_o$, learning rate $\eta$, number of training steps $K$, batch size $B$.
\State \textbf{Output:} Updated model parameters $\theta$.
\For{$k = 1$ to $K$}
    \State Acquire retain and forget mini-batches $D_r, D_f$  of size $B$ from $\mathcal{D}_r, \mathcal{D}_f$ respectively.
    \State  $L_r \gets \frac{1}{B}\sum_{x\in D_r}\mathcal{L}(\mathcal M_\theta, x)$
    \State $\gr \gets \nabla_\theta L_r$
    \State $\gf \gets \frac{1}{B}\sum_{x\in D_f}\nabla_{\theta} \mathcal{L}(\mathcal M_{\theta}, x)$
    \State $L \gets L_r + \beta_o \left(\frac{\gr \cdot \gf}{\|\gr\|\|\gf\|}\right)^2$
    
    \State $\theta \gets \theta - \eta \nabla_\theta L$
\EndFor
\State \Return $\theta$
\end{algorithmic}
\end{algorithm}
%%%%%%%%%%%%%%%%%%%%%%%%%%%

%%%%%%%%%%%%%%%%%%%%%%%%%%%
\begin{algorithm}[!htp]
\caption{Alternating orthogonalization and surgery (UNO-S)}\label{algo:UNO-S}
\begin{algorithmic}[1]
\State \textbf{Input:} Loss function $\mathcal{L}$, retain dataset $\mathcal{D}_r$, forget dataset $\mathcal{D}_f$, trained model requiring unlearning $\mathcal{M}_\theta$, weight for orthogonalization loss term $\beta_o$, learning rate $\eta$, number of training steps $K$, batch size $B$.
\State \textbf{Output:} Updated model parameters $\theta$.
\For{$k = 1$ to $K$}
    \State Acquire retain and forget mini-batches $D_r, D_f$  of size $B$ from $\mathcal{D}_r, \mathcal{D}_f$ respectively.
    \State  $L_r \gets \frac{1}{B}\sum_{x\in D_r}\mathcal{L}(\mathcal M_\theta, x)$
    \State $\gr \gets \nabla_\theta L_r$
    \State $\gf \gets \frac{1}{B}\sum_{x\in D_f}\nabla_{\theta} \mathcal{L}(\mathcal M_{\theta}, x)$
    \If{$k$ is odd}
    \State $L \gets L_r + \beta_o \left( \frac{\gr \cdot \gf}{\|\gr\| \|\gf\|} \right)^2$
    \State $\theta \gets \theta - \eta \nabla_\theta L$
    \Else
    \State $\gr \gets \gr - \frac{\gr \cdot \gf}{\|\gf\|^2}\gf$
    \State $\theta \gets \theta - \eta  \gr$
    \EndIf
\EndFor
\State \Return $\theta$
\end{algorithmic}
\end{algorithm}
%%%%%%%%%%%%%%%%%%%%%%%%%%%

%%%%%%%%%%%%%%%%%%%%%%%%%%%%%%%%%%%%%%%

%%%%%%%%%%%%%%%%%%%%%%%%%%%
\begin{algorithm}[!htp]
\caption{Gradient surgery with histogram unlearning (\shat)}\label{algo:Sh}
\begin{algorithmic}[1]
\State \textbf{Input:} Loss function $\mathcal{L}$, retain dataset $\mathcal{D}_r$, forget dataset $\mathcal{D}_f$, trained model requiring unlearning $\mathcal{M}_\theta$, learning rate $\eta$, number of training steps $K$, batch size $B$, number of samples to generate $N_g$, classifier model $\mathcal C_\phi$, weight for KL divergence loss term $\beta_h$, a small positive threshold for stabilizing KL divergence computation $\alpha$.
\State \textbf{Output:} Updated model parameters $\theta$.
\For{$k = 1$ to $K$}
    \State Acquire retain and forget mini-batches $D_r, D_f$  of size $B$ from $\mathcal{D}_r, \mathcal{D}_f$ respectively.
    \State Generate $N_g$ samples $\{y_i\}_{i=1}^{N_g}$ using $\mathcal M_\theta$.
    \State $p_r \gets \frac{1}{N_g}\sum_{i=1}^{N_g}\mathcal C_\phi(y_i)$
    \State $d_{\rm KL} \gets p_r \log\left(\frac{p_r}{1 - \alpha}\right) + (1 - p_r) \log\left(\frac{1 - p_r}{\alpha}\right)$
    \State  $\gr \gets \frac{1}{B}\sum_{x\in D_r}\nabla_\theta \mathcal{L}(\mathcal M_\theta, x) + \beta_h \nabla_\theta d_{\rm KL}$
    \State $\gf \gets \frac{1}{B}\sum_{x\in D_f}\nabla_{\theta} \mathcal{L}(\mathcal M_{\theta}, x)+ \beta_h \nabla_\theta d_{\rm KL}$
    \State $\gr \gets \gr - \frac{\gr \cdot \gf}{\|\gf\|^2}\gf$
    \State $\theta \gets \theta - \eta  \gr$
\EndFor
\State \Return $\theta$
\end{algorithmic}
\end{algorithm}
%%%%%%%%%%%%%%%%%%%%%%%%%%%

%%%%%%%%%%%%%%%%%%%%%%%%%%%
\begin{algorithm}[!htp]
\caption{UNO with histogram unlearning (\unoh)}\label{algo:UNOh}
\begin{algorithmic}[1]
\State \textbf{Input:} Loss function $\mathcal{L}$, retain dataset $\mathcal{D}_r$, forget dataset $\mathcal{D}_f$, trained model requiring unlearning $\mathcal{M}_\theta$, weight for orthogonalization loss term $\beta_o$, learning rate $\eta$, number of training steps $K$, batch size $B$, number of samples to generate $N_g$, classifier model $\mathcal C_\phi$, weight for KL divergence loss term $\beta_h$, a small positive threshold for stabilizing KL divergence computation $\alpha$.
\State \textbf{Output:} Updated model parameters $\theta$.
\For{$k = 1$ to $K$}
    \State Acquire retain and forget mini-batches $D_r, D_f$  of size $B$ from $\mathcal{D}_r, \mathcal{D}_f$ respectively.
    \State Generate $N_g$ samples $\{y_i\}_{i=1}^{N_g}$ using $\mathcal M_\theta$.
    \State $p_r \gets \frac{1}{N_g}\sum_{i=1}^{N_g}\mathcal C_\phi(y_i)$
    \State $d_{\rm KL} \gets p_r \log\left(\frac{p_r}{1 - \alpha}\right) + (1 - p_r) \log\left(\frac{1 - p_r}{\alpha}\right)$
    \State  $L_r \gets \frac{1}{B}\sum_{x\in D_r}\mathcal{L}(\mathcal M_\theta, x) + \beta_h d_{\rm KL}$
    \State $\gr \gets \nabla_\theta L_r$
    \State $\gf \gets \frac{1}{B}\sum_{x\in D_f}\nabla_{\theta} \mathcal{L}(\mathcal M_{\theta}, x)+ \beta_h \nabla_\theta d_{\rm KL}$
    \State $L \gets L_r + \beta_o \left(\frac{\gr \cdot \gf}{\|\gr\|\|\gf\|}\right)^2$
    
    \State $\theta \gets \theta - \eta \nabla_\theta L$
\EndFor
\State \Return $\theta$
\end{algorithmic}
\end{algorithm}
%%%%%%%%%%%%%%%%%%%%%%%%%%%

%%%%%%%%%%%%%%%%%%%%%%%%%%%
\begin{algorithm}[!htp]
\caption{Alternating orthogonalization and surgery with histogram unlearning (\unosh)}\label{algo:UNO-Sh}
\begin{algorithmic}[1]
\State \textbf{Input:} Loss function $\mathcal{L}$, retain dataset $\mathcal{D}_r$, forget dataset $\mathcal{D}_f$, trained model requiring unlearning $\mathcal{M}_\theta$, weight for orthogonalization loss term $\beta_o$, learning rate $\eta$, number of training steps $K$, batch size $B$, number of samples to generate $N_g$, classifier model $\mathcal C_\phi$, weight for KL divergence loss term $\beta_h$, a small positive threshold for stabilizing KL divergence computation $\alpha$.
\State \textbf{Output:} Updated model parameters $\theta$.
\For{$k = 1$ to $K$}
    \State Acquire retain and forget mini-batches $D_r, D_f$  of size $B$ from $\mathcal{D}_r, \mathcal{D}_f$ respectively.
    \State Generate $N_g$ samples $\{y_i\}_{i=1}^{N_g}$ using $\mathcal M_\theta$.
    \State $p_r \gets \frac{1}{N_g}\sum_{i=1}^{N_g}\mathcal C_\phi(y_i)$
    \State $d_{\rm KL} \gets p_r \log\left(\frac{p_r}{1 - \alpha}\right) + (1 - p_r) \log\left(\frac{1 - p_r}{\alpha}\right)$
    \State  $L_r \gets \frac{1}{B}\sum_{x\in D_r}\mathcal{L}(\mathcal M_\theta, x) + \beta_h d_{\rm KL}$
    \State $\gr \gets \nabla_\theta L_r$
    \State $\gf \gets \frac{1}{B}\sum_{x\in D_f}\nabla_{\theta} \mathcal{L}(\mathcal M_{\theta}, x)+ \beta_h \nabla_\theta d_{\rm KL}$
    \If{$k$ is odd}
    \State $L \gets L_r + \beta_o \left(\frac{\gr \cdot \gf}{\|\gr\|\|\gf\|}\right)^2$
    \State $\theta \gets \theta - \eta \nabla_\theta L$
    \Else
    \State $\gr \gets \gr - \frac{\gr \cdot \gf}{\|\gf\|^2}\gf$
    \State $\theta \gets \theta - \eta  \gr$
    \EndIf
\EndFor
\State \Return $\theta$
\end{algorithmic}
\end{algorithm}
%%%%%%%%%%%%%%%%%%%%%%%%%%%

%%%%%%%%%%%%%%%%%%%%%%%%%%%
\begin{algorithm}[!htp]
\caption{Histogram unlearning (H)}\label{algo:H}
\begin{algorithmic}[1]
\State \textbf{Input:} Loss function $\mathcal{L}$, retain dataset $\mathcal{D}_r$, forget dataset $\mathcal{D}_f$, trained model requiring unlearning $\mathcal{M}_\theta$, learning rate $\eta$, number of training steps $K$, batch size $B$, number of samples to generate $N_g$, classifier model $\mathcal C_\phi$, weight for KL divergence loss term $\beta_h$, a small positive threshold for stabilizing KL divergence computation $\alpha$.
\State \textbf{Output:} Updated model parameters $\theta$.
\For{$k = 1$ to $K$}
    \State Acquire retain and forget mini-batches $D_r, D_f$  of size $B$ from $\mathcal{D}_r, \mathcal{D}_f$ respectively.
    \State Generate $N_g$ samples $\{y_i\}_{i=1}^{N_g}$ using $\mathcal M_\theta$.
    \State $p_r \gets \frac{1}{N_g}\sum_{i=1}^{N_g}\mathcal C_\phi(y_i)$
    \State $d_{\rm KL} \gets p_r \log\left(\frac{p_r}{1 - \alpha}\right) + (1 - p_r) \log\left(\frac{1 - p_r}{\alpha}\right)$
    \State  $L \gets \frac{1}{B}\sum_{x\in D_r}\mathcal{L}(\mathcal M_\theta, x) + \beta_h d_{\rm KL}$
    \State $\theta \gets \theta - \eta \nabla_\theta L$
\EndFor
\State \Return $\theta$
\end{algorithmic}
\end{algorithm}
%%%%%%%%%%%%%%%%%%%%%%%%%%%
%%%%%%%%%%%%%%%%%%%%%%%%%%%
\subsection{Unlearning algorithms for benchmarking}
In this section we provide the pseudocode for the unlearning algorithms we use for benchmarking in Table~\ref{tab:res}. Algorithm~\ref{algo:vae_fu} presents the latent-space feature unlearning method for VAEs proposed by \citet{moon2024feature}, while Algorithm~\ref{algo:gandikota} describes the concept erasure method for diffusion transformers introduced by \citet{gandikota2023erasing}.
% Finally, Algorithm~\ref{algo:salun} outlines the saliency-based unlearning method for diffusion models proposed by \citet{fan2024salun}.

\begin{algorithm}[!htp]
\caption{Feature unlearning for VAEs~\citep{moon2024feature}}\label{algo:vae_fu}
\begin{algorithmic}[1]
\State \textbf{Input:} Retain dataset $\mathcal{D}_r$, forget dataset $\mathcal{D}_f$, trained VAE requiring unlearning $\mathcal{M}_\theta$, latent dimension $d$, learning rate $\eta$, number of training steps $K$, batch size $B$, unlearning weight $\alpha_{\rm FU}$.
\State \textbf{Output:} Updated VAE parameters $\theta$.

\State Create a frozen reference VAE:
$
    \mathcal{M}_{\theta_0} \gets \mathrm{Freeze}(\mathrm{Copy}(\mathcal{M}_\theta)).
$

\State Collect latent means from the frozen encoder:
$
    Z_f \gets \{\mu_{\theta_0}(x): x \in \mathcal{D}_f\},
    \;
    Z_r \gets \{\mu_{\theta_0}(x): x \in \mathcal{D}_r\}.
$

\State Compute the forget and retain latent centers:
$
    \mu_f \gets \frac{1}{|Z_f|}\sum_{z\in Z_f} z,
    \;
    \mu_r \gets \frac{1}{|Z_r|}\sum_{z\in Z_r} z.
$

\State Compute the latent feature direction:
$
    z_e \gets \mu_f - \mu_r,
    \;
    v \gets \frac{z_e}{\|z_e\|}.
$

\State Compute scalar projections:
$a_f \gets \frac{1}{|Z_f|}\sum_{z\in Z_f} z^\top v,\;
    a_r \gets \frac{1}{|Z_r|}\sum_{z\in Z_r} z^\top v.$

\State Set the threshold:
$
    \delta \gets \frac{1}{2}(a_f+a_r).
$

\For{$k = 1$ to $K$}
    \State Acquire mini-batches $D_r$ and $D_f$ of size $B$.
    \State Sample random latent vectors:
    $
        Z \gets \{z_i\}_{i=1}^{B},
        \qquad
        z_i \sim \mathcal{N}(0,I_d).
    $

    \For{each $z_i \in Z$}
        \State Compute the forget-feature score:
        $
            p_i \gets z_i^\top v.
        $

        \State Compute the forget indicator:
        $
            s_i \gets \mathbf{1}\{p_i > \delta\}.
        $

        \State Remove the excess forget feature:
        $
            \hat z_i
            \gets
            z_i - \max(p_i-\delta,0)z_e.
        $
    \EndFor

    \State Decode the original and edited latents:
    \State$
        g_i \gets \mathcal{M}_{\theta}.\mathrm{decode}(z_i),\;
        f_i \gets \mathcal{M}_{\theta_0}.\mathrm{decode}(z_i),\;
        \hat f_i \gets \mathcal{M}_{\theta_0}.\mathrm{decode}(\hat z_i).
    $
   \State Compute losses:
   \State
    $
        \mathcal{L}_{\mathrm{recon}}
        \gets
        \frac{1}{B}
        \sum_{i=1}^{B}
        (1-s_i)\|g_i-f_i\|_1.
    $

    \State 
    $
        \mathcal{L}_{\mathrm{unlearn}}
        \gets
        \frac{1}{B}
        \sum_{i=1}^{B}
        s_i\|g_i-\hat f_i\|_1.
    $

    \State 
    $
        \mathcal{L}_{\mathrm{percep}}
        \gets
        \frac{1}{B}
        \sum_{i=1}^{B}
        s_i\left(1-\mathrm{SSIM}(g_i,\hat f_i)\right).
    $

    \State Compute the full objective:
    $
        \mathcal{L}_{\mathrm{FU}}
        \gets
        \mathcal{L}_{\mathrm{recon}}
        +
        \alpha_{\rm FU}
        \left(
        \mathcal{L}_{\mathrm{unlearn}}
        +
        \mathcal{L}_{\mathrm{percep}}
        \right).
    $

    \State Update VAE parameters by gradient descent:
    $
        \theta \gets \theta - \eta \nabla_\theta \mathcal{L}_{\mathrm{FU}}.
    $
\EndFor

\State \Return $\theta$
\end{algorithmic}
\end{algorithm}
%%%%%%%%%%%%%%%%%%%%%%%%%%%
%%%%%%%%%%%%%%%%%%%%%%%%%%%
\begin{algorithm}[!htp]
\caption{Concept erasure for diffusion models~\citep{gandikota2023erasing}}\label{algo:gandikota}
\begin{algorithmic}[1]
\State \textbf{Input:} Forget dataset $\mathcal{D}_f$, trained model requiring unlearning $\mathcal{M}_\theta$, VAE encoder $\mathcal{E}$, diffusion schedule $\{\bar\alpha_t\}_{t=0}^{T}$, learning rate $\eta$, erasure strength $\lambda$, number of training steps $K$, batch size $B$.
\State \textbf{Output:} Updated model parameters $\theta$.
\State Create a frozen teacher model $\mathcal{T}_{\theta_0} \gets \mathrm{Freeze}(\mathrm{Copy}(\mathcal{M}_\theta))$.
\For{$k = 1$ to $K$}
    \State Acquire mini-batch $D_f=\{(x_i,y_i)\}_{i=1}^{B}$ from $\mathcal{D}_f$.
    \State Sample diffusion timesteps $t_i \sim \mathrm{Uniform}\{0,\ldots,T\}$.
    \State Encode images into latent space:
    $
        z_i \gets \mathcal{E}(x_i).
    $
    \State Sample noise $\epsilon_i \sim \mathcal{N}(0,I)$.
    \State Construct noisy latents:
    $
        x_{t_i}^{(i)} \gets \sqrt{\bar\alpha_{t_i}}z_i 
        + \sqrt{1-\bar\alpha_{t_i}}\epsilon_i.
    $
    \State Let $y_{\emptyset}$ denote the null class label.
    \State Compute the erased teacher target:
    $
        \epsilon_{\mathrm{target}}^{(i)}
        \gets
        (1+\lambda)\mathcal{T}_{\theta_0}(x_{t_i}^{(i)},t_i,y_{\emptyset})
        -
        \lambda \mathcal{T}_{\theta_0}(x_{t_i}^{(i)},t_i,y_i).
    $
    \State Compute the model prediction:
    $
        \epsilon_{\theta}^{(i)}
        \gets
        \mathcal{M}_{\theta}(x_{t_i}^{(i)},t_i,y_i).
    $
    \State Compute the erasure loss:
    $
        \mathcal{L}_{\mathrm{erase}}
        \gets
        \frac{1}{B}\sum_{i=1}^{B}
        \left\|
        \epsilon_{\theta}^{(i)}
        -
        \epsilon_{\mathrm{target}}^{(i)}
        \right\|^2.
    $
    \State Update model parameters by gradient descent:
    $
        \theta \gets \theta - \eta \nabla_{\theta}\mathcal{L}_{\mathrm{erase}}.
    $
\EndFor
\State \Return $\theta$
\end{algorithmic}
\end{algorithm}
\newpage
\section{Catastrophic forgetting induced by gradient ascent and ascent–descent}\label{app:catastrophy}
Figure~\ref{fig:catastrophic} shows the generated samples for unlearning the digit $1$ after $49$ parameter update steps of gradient ascent~\eqref{eq:A} and ascent-descent~\eqref{eq:A-D} at a learning rate of $\eta=10^{-3}$. In this example, the $1$’s were successfully forgotten, but all other digits were forgotten as well. In particular, the left panel of Figure~\ref{fig:catastrophic} shows that the model only remembers the complement of $1$'s after gradient ascent.

\begin{figure}[h]
    \centering
    \includegraphics[width=0.4\textwidth]{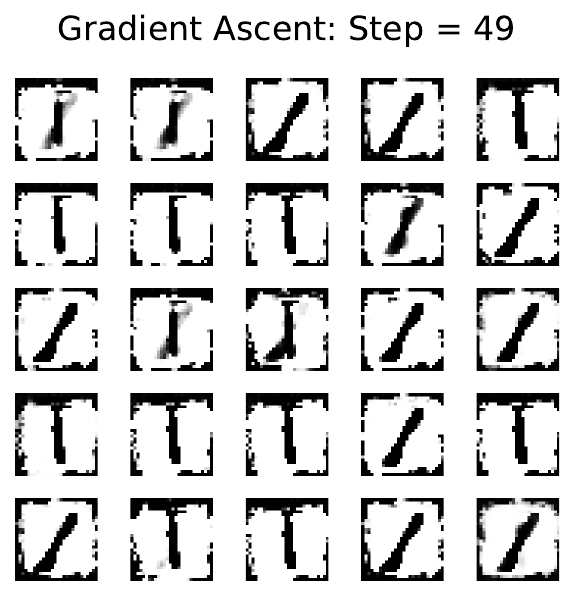}
    \hspace{0.02\textwidth}
    \includegraphics[width=0.4\textwidth]{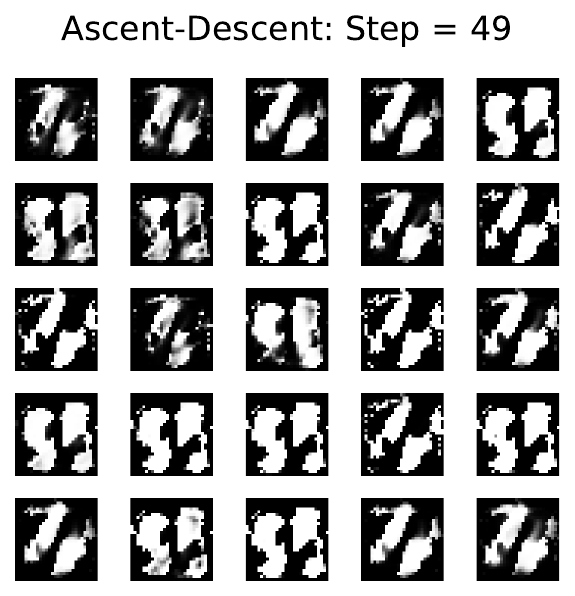}
    \caption{Catastrophic forgetting induced by gradient ascent (left) and ascent-descent (right) for MNIST at a learning rate of $10^{-3}$.}
    \label{fig:catastrophic}
\end{figure}

%%%%%%%%%%%%%%%%%%%%%%%%%%%%%%%%%%%%%%%%%%%%%%%%%%%%%%%%%%%%%%%%%%%%%%%%%%%%%%
\newpage
\section{Comparison of two variants of gradient surgery}\label{app:gs}
We now compare two variants of gradient surgery: 1) gradient surgery with descent in the retain direction (S), described in Algorithm~\ref{algo:S}, used throughout this paper and 2) gradient surgery with ascent in the forget direction (SA), described in Algorithm~\ref{algo:SA} which appears in \citet{bae2023gradient}. Figure~\ref{fig:sa-lr} shows that SA is prone to catastrophic forgetting and requires a carefully tuned, small learning rate to mitigate this effect. But even with a small learning rate the generated samples might look significantly different from the original model; for samples generated by the original model, see Figure~\ref{fig:mnist-transform}. On the other hand, Figure~\ref{fig:s-lr} shows that S does not suffer from catastrophic forgetting, even for a large learning rate applied for many training steps, and produces samples that are much closer to the original model.

\begin{figure}[h]
    \centering
\includegraphics[width=\linewidth]{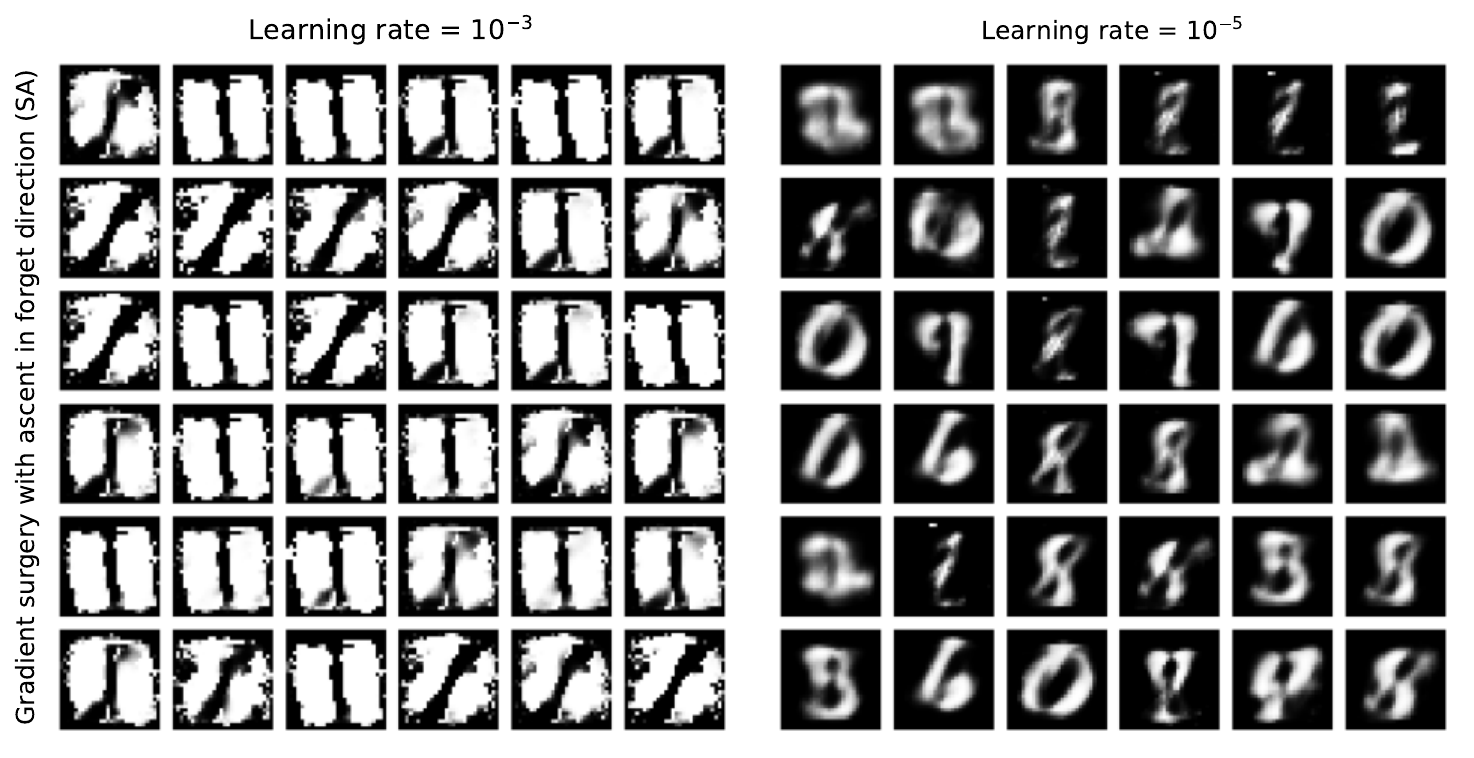}
    \caption{Generated samples after unlearning digit $1$ via gradient surgery with ascent in the forget direction (SA), described in Algorithm~\ref{algo:SA}, for two different learning rates: $10^{-3}$ (left), $10^{-5}$ (right). SA was run for $K=53$ training steps on the left and $K=530$ training steps on the right.}
    \label{fig:sa-lr}
\end{figure}

\begin{figure}[h]
    \centering
\includegraphics[width=\linewidth]{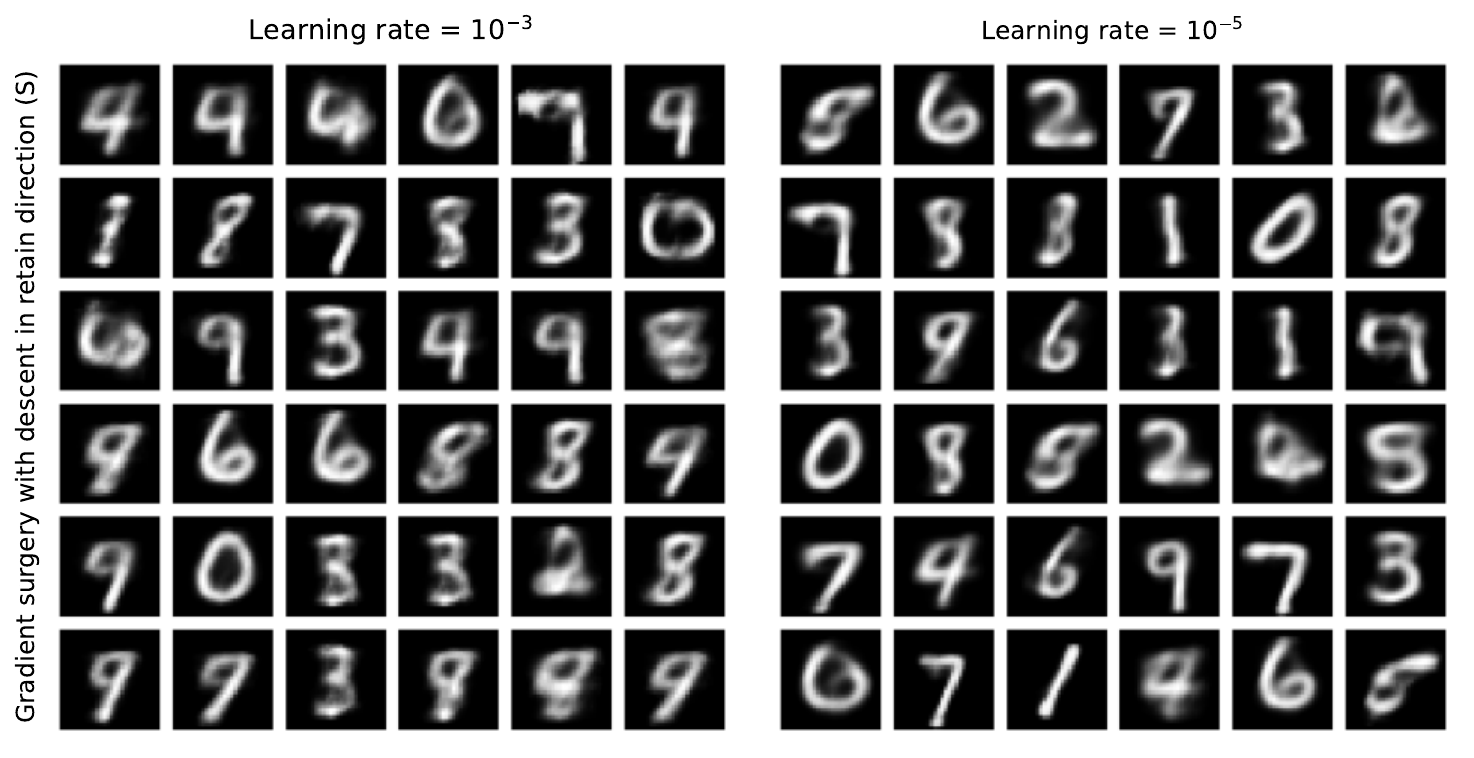}
    \caption{Generated samples after unlearning digit $1$ via gradient surgery with descent in the retain direction (S), described in Algorithm~\ref{algo:S}, for two different learning rates: $10^{-3}$ (left), $10^{-5}$ (right). S was run for $K=530$ training steps for both learning rates.}
    \label{fig:s-lr}
\end{figure}

%%%%%%%%%%%%%%%%%%%%%%%%%%%%%%%%%%%%%%%%%%%%%%%%%%%%%%%%%%%%%%%%%%%%%%%%%%%%%%

\section{Additional samples and experiments for CelebA}
%before and after unlearning}
\label{app:celeba-more}
Figure~\ref{fig:celeba-more} shows $18$ pairs of images generated before and after unlearning with UNO for CelebA. In many of these pairs the after image shows a subtly larger smile than the before image, see for example the fourteenth pair. This is consistent with the phenomenon that women tend to smile more than men in photographs~\citep{wondergem2012gender}.

\begin{figure}[h]
    \centering
\includegraphics[width=\linewidth]{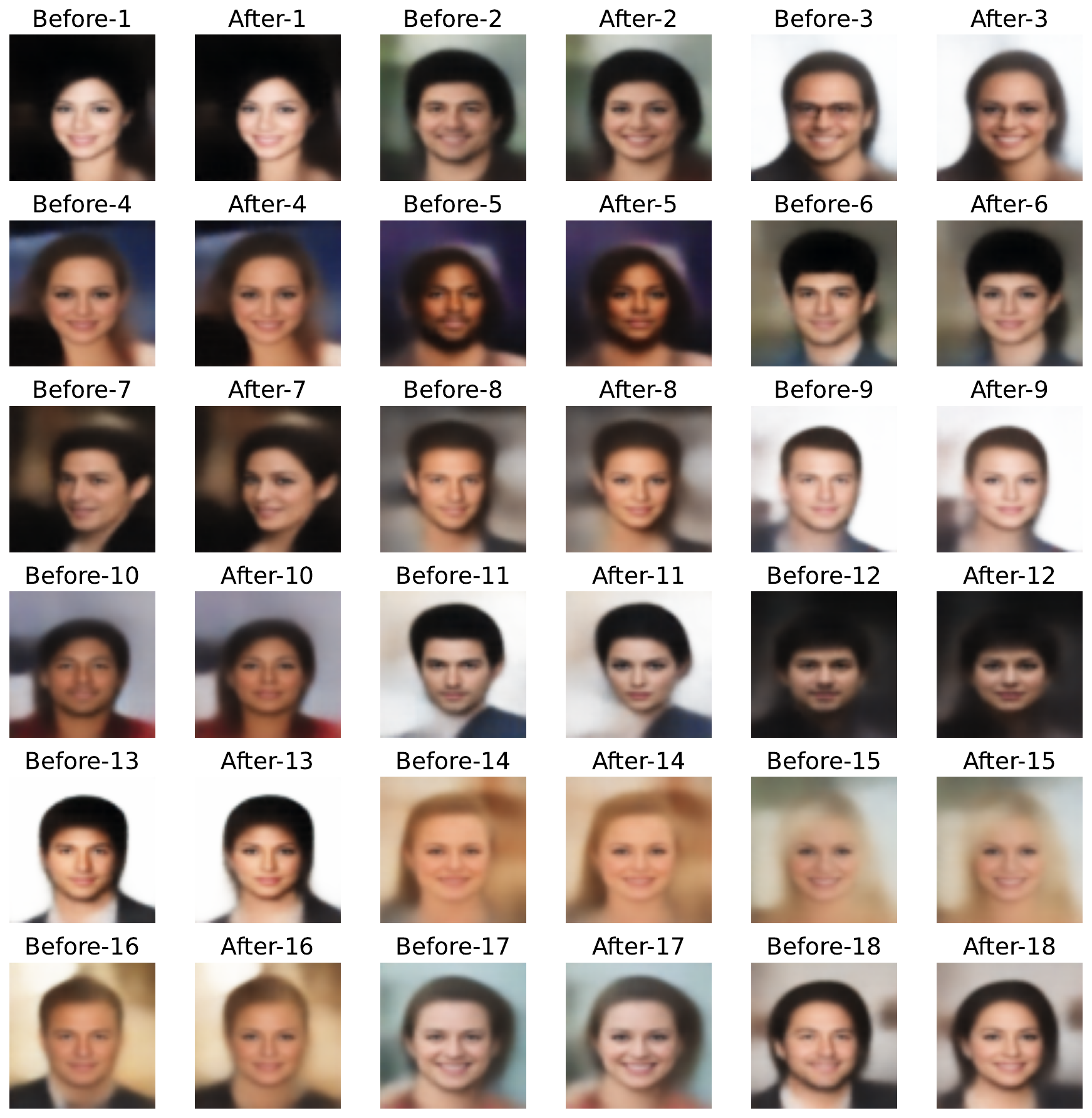}
    \caption{Results for unlearning on CelebA with UNO, illustrated using 18 pairs of generated images. The images labeled "Before" were generated using the original model. Each image labeled "After" was generated  after unlearning using the same noise sample for the decoder as the corresponding "Before" image.}
    \label{fig:celeba-more}
\end{figure}

\subsection{Eyeglasses removal with unlearning} 
\label{app:eyeglasses} 
Out of the $202,599$ images in CelebA, $13,193$ or roughly $6.5\%$ contain faces with eyeglasses. By treating the images with eyeglasses as the forget set and those without eyeglasses as the retain set, we can apply our unlearning algorithms to remove the presence of eyeglasses from the generated samples. Figure~\ref{fig:glasses-large-gen} shows $18$ pairs of generated samples before and after unlearning with UNO-S for the same noise samples used for the decoder. For some of these before–after pairs, where the before image contains opaque, dark eyeglasses, the after image may exhibit darker regions around the eyes, resembling periorbital hyperpigmentation~\citep{sarkar2016periorbital} (see, for example, the last pair, labelled 18). This phenomenon does not occur for images with more transparent eyeglasses. During the original training instance, the model likely conflates the concept of dark eyewear with hyperpigmentation to some extent, due to its finite resolution capabilities. A larger model, capable of learning finer-grained patterns, might better distinguish between these two concepts, and therefore may not produce this phenomenon after unlearning. The hyperparameter values used in these experiments are identical to those reported in Table~\ref{tab:param}.

\begin{figure}[h]
    \centering
\includegraphics[width=\linewidth]{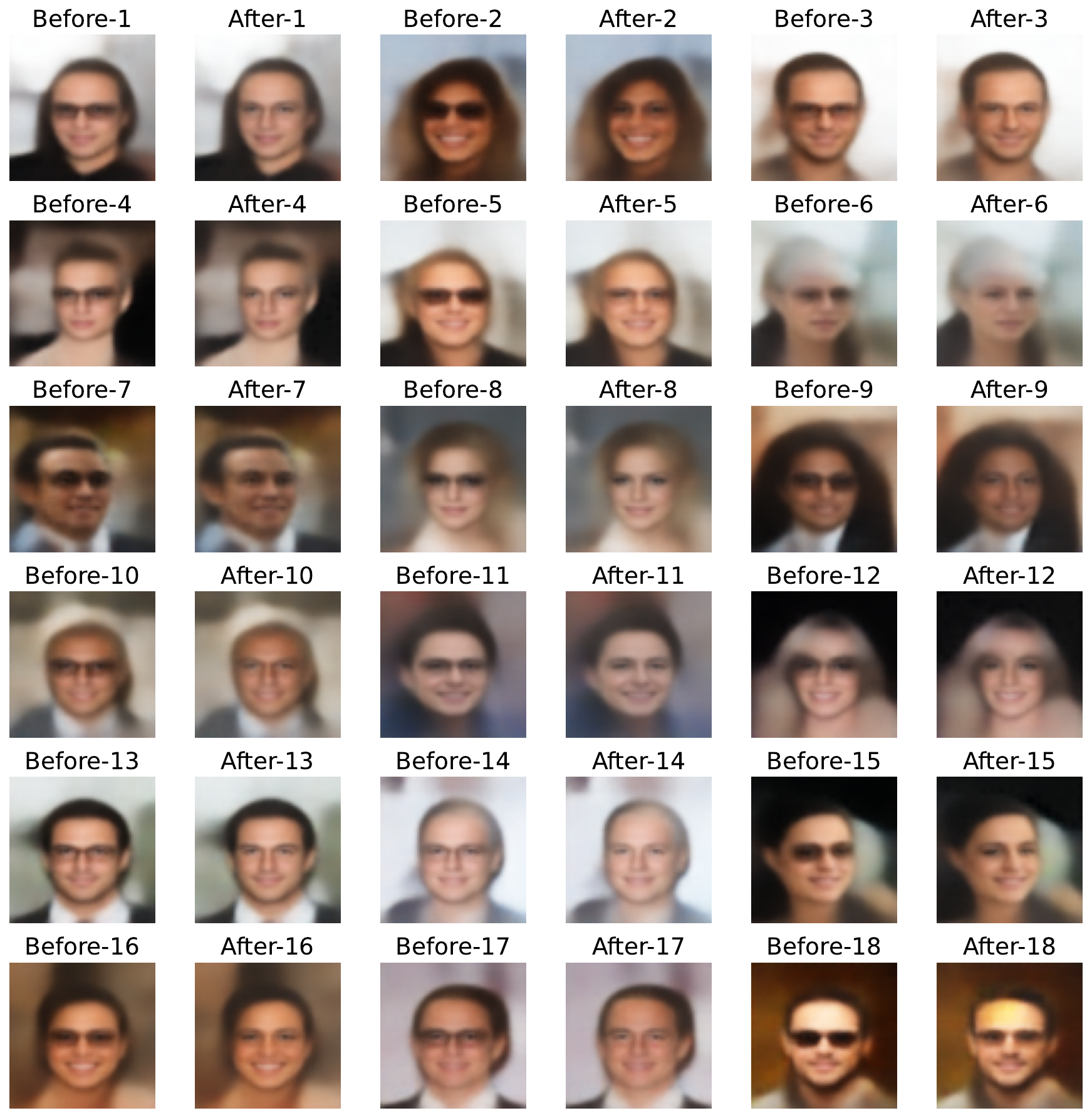}
    \caption{Results for eyeglasses removal on CelebA with UNO-S, illustrated using 18 pairs of generated images. The images labeled "Before" were generated using the original model. Each image labeled "After" was generated after unlearning using the same noise sample for the decoder as the corresponding "Before" image.}
    \label{fig:glasses-large-gen}
\end{figure}

{
\subsection{Nonsequential multiclass unlearning} 
\label{app:multiclass} 
In our previous experiments we unlearned a single class, e.g. "male" or "eyeglasses". We consider now the important application of multiclass unlearning. This situation is more complex than single class unlearning as the different forgetting objectives typically interact through the shared parameters of the model. For example, unlearning information about a class $c_i$ may undo forgetting of the class $c_j$ and vice versa. Hence, multiclass unlearning can be viewed as an multi-objective optimization problem.

In this section we show results for multiclass unlearning where we forget samples that are labelled as "male" and having dark hair i.e. either "brown" or "black" hair. This can be thought of as a composite class and in terms of the labelling of the dataset can be represented as ((Male = 1) \& ((Black\_Hair = 1) | (Brown\_Hair = 1))). In Figure~\ref{fig:celeba-multi} we see that images produced by the original model with male and dark-hair attributes get converted into images of females with dark hair. Generated images of men wearing hats or having blond hair and images of females remain roughly unaltered after unlearning. It is interesting to note that unlearning does not change the hair color attribute but rather changes the "male" attribute, indicating that while unlearning a composite class, certain features are forgotten while other shared features are kept. 

\begin{figure}[h]
    \centering
    \includegraphics[width=\linewidth]{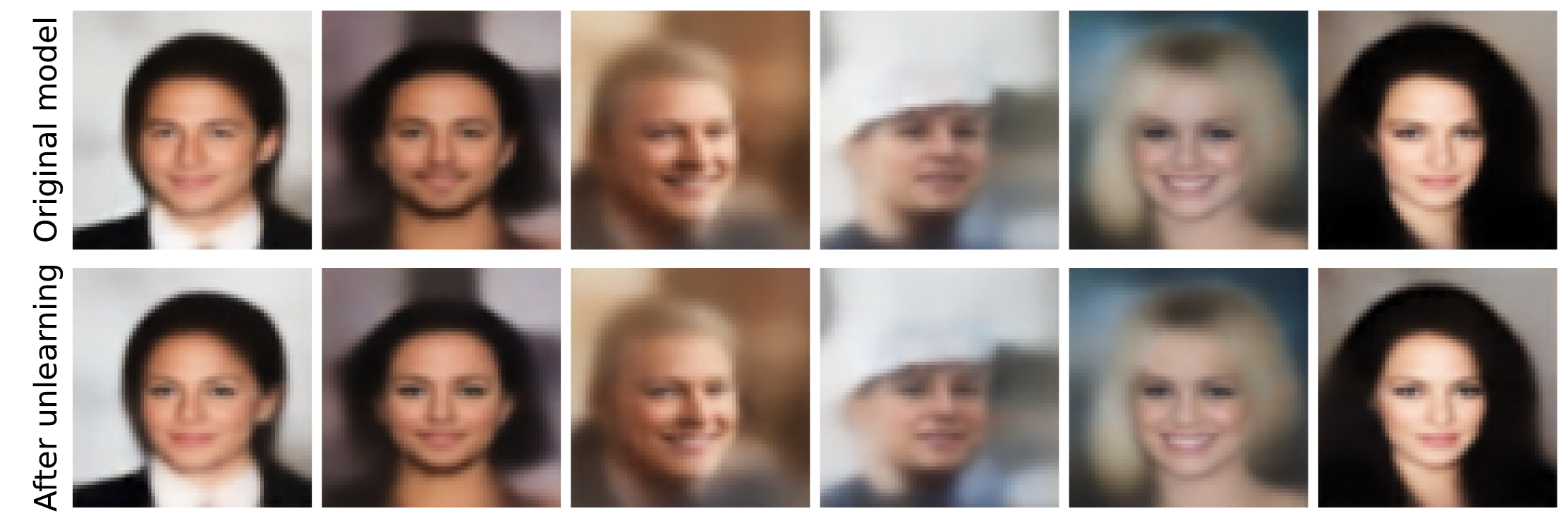}
    \caption{Results for multiclass unlearning on CelebA with UNO-S, where the composite class "male with dark hair" i.e. ((Male = 1) \& ((Black\_Hair = 1) | (Brown\_Hair = 1))) is unlearned. The same noise samples were decoded by the original model (top) and the unlearned model (bottom) to generate the faces.}
    \label{fig:celeba-multi}
\end{figure}

}
%%%%%%%%%%%%%%%%%%%%%%%%%%%%%%%%%%%%%%%%%%%%%%%%%%%%%%%%%%%%%%%%%%%%%%%%%%%%%%

\section{Additional samples for ImageNet-1K}
% before and after unlearning}
\label{app:imagenet-more}
Figure~\ref{fig:imagenet-more} shows $18$ pairs of images generated before and after unlearning with UNO-S for ImageNet-1K. The images exhibit a significant degree of variation, however, they clearly remain in the same class as identified by the pretrained classifier.

\begin{figure}[h]
    \centering
\includegraphics[width=\linewidth]{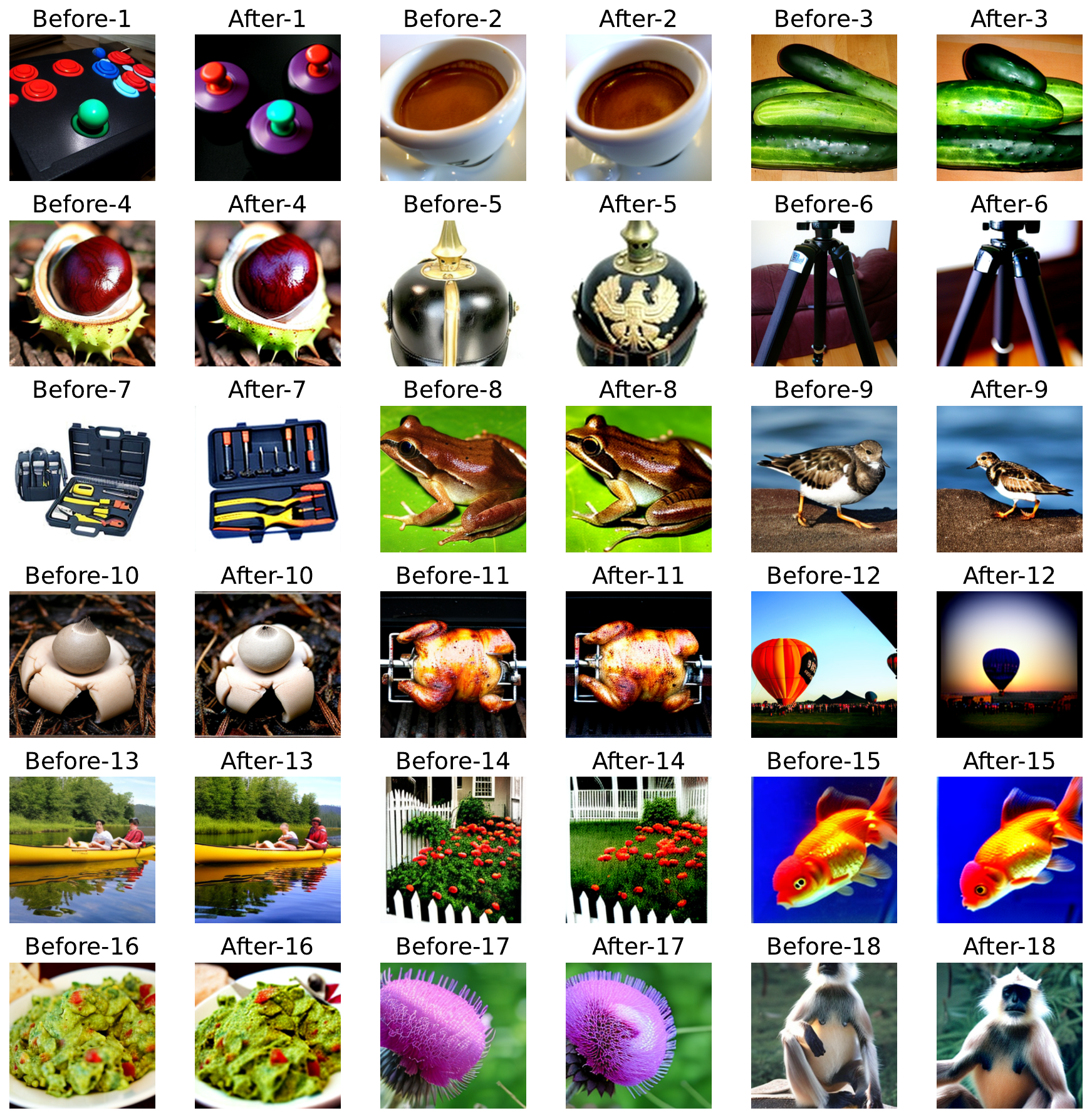}
    \caption{Results for unlearning on ImageNet with UNO-S, illustrated using 18 pairs of generated images. The images labeled "Before" were generated using the original model. Each image labeled "After" was generated  after unlearning using the same noise sample for the diffusion transformer as the corresponding "Before" image.}
    \label{fig:imagenet-more}
\end{figure}

{Figure~\ref{fig:imagenet-ratios-UNO}, we show how UNO and gradient surgery are able to perform successful replacement learning for ratios of the cardinalities of the retain dataset and the forget dataset, respectively. We show results for both, replacement unlearning where the classes share many features (golden retrievers to labrador retrievers) and where they do not share many common features (golden retrievers to mopeds). UNO and gradient surgery perform both equally well as UNO-S (cf. Figure~\ref{fig:imagenet-ratios}).}

\begin{figure}[!htp]
    \centering
    \includegraphics[width=\linewidth]{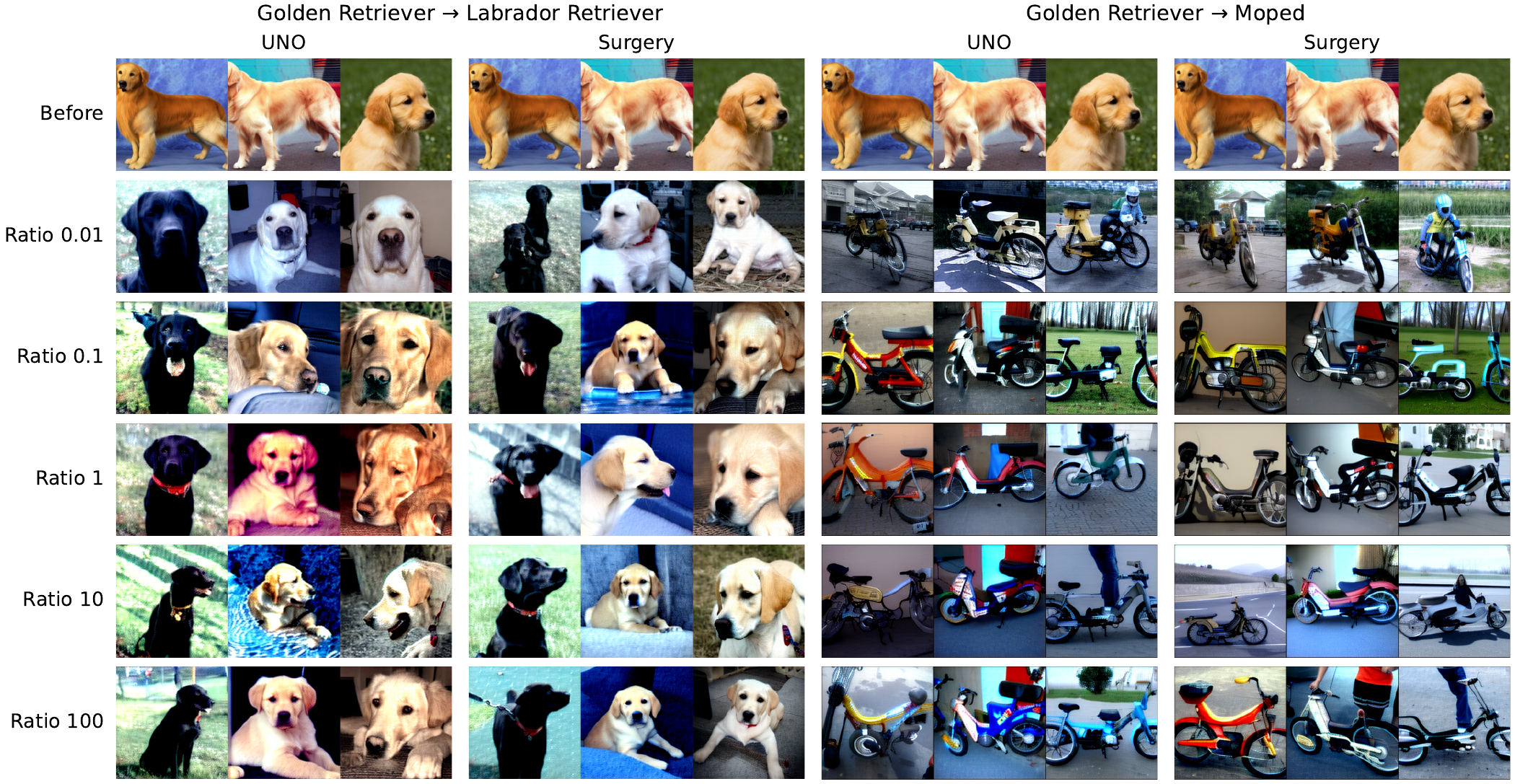}
    \caption{{ImageNet-1K samples generated by the original model (top) and after replacement unlearning with UNO for different ratios of the cardinalities of the retain and forget datasets. We show replacement unlearning of class $207$ (golden retriever) with class $208$ (labrador retriever) (left) and replacement unlearning of class $207$ (golden retriever) with class $665$ (moped) (right). We use identical noise samples for generating the before and after images.}}
    \label{fig:imagenet-ratios-UNO}
\end{figure}

%%%%%%%%%%%%%%%%%%%%%%%%%%%%%%%%%%%%%%%%%%%%%%%%%%%%%%%%%%%%%%%%%%%%%%%%%%%%%%

\section{Results for classifier-assisted unlearning}
\label{app:un-w-cls}
In this section, we present the results for the algorithms introduced in Section~\ref{ssec:hat}. Comparing Table~\ref{tab:res} with Table~\ref{tab:res-hat} shows that \shat~achieves orders of magnitude speed-up over S for both MNIST and CelebA. UNO and UNO-S, already fast, do not gain significant speed-up with classifier-assistance. Histogram unlearning (H), although successful, is much slower than the other successful algorithms in Table~\ref{tab:res-hat}. All algorithms in Table~\ref{tab:res-hat} preserve the fidelity of the original model, with \unoh~producing the lowest FID for both datasets.  
\begingroup
\begin{table}[h]
\caption{Performance of various algorithms for class/feature unlearning with VAEs on MNIST and CelebA when a classifier able to distinguish between the retain and forget data is available. Each experiment is repeated $10$ times, and the standard deviations are shown in parentheses. Bold indicates the best score. {\color{red}\ding{55}} indicates that the generated samples after unlearning are unrecognizably different from the original model. {\color{green!60!black}\ding{51}} indicates the generated samples after unlearning are perceptually indistinguishable from the original model in terms of visual fidelity. An asterisk (*) indicates that, without classifier assistance, the algorithm failed to reach the target fraction of forget samples in the generated images within the allotted training steps.}
\label{tab:res-hat}
\centering
\begin{tabular}{@{}llcccc@{}}
\toprule
\textbf{\qquad Dataset} & \textbf{\hspace{-0.5em}Algorithm} & Time to unlearn (s) $\downarrow$ & FID $\downarrow$ & Time per step (s) \\
\midrule
\multirow{3}{*}{\raisebox{-2.8ex}[0pt][0pt]{\shortstack[c]{MNIST \\ (Class: 1)\\Original FID: 20.7}}}
& H   & 2.181 (0.853)         & 23.4 (0.5) {\color{green!60!black}\ding{51}}         & 0.006 (0.0003)   \\
  & \shat              & \textbf{0.021} (0.003)         & 24.1 (0.5) {\color{green!60!black}\ding{51}} & 0.008 (0.0006)   \\
  & \unoh            & 0.061 (0.009)         & \textbf{22.1} (0.5) {\color{green!60!black}\ding{51}}          & 0.022 (0.0002)   \\
  & \unosh  & 0.041 (0.021)& \textbf{22.1} (0.4) {\color{green!60!black}\ding{51}}          & 0.017 (0.0067)   \\
\midrule
\multirow{3}{*}{\raisebox{-2.8ex}[0pt][0pt]{\shortstack[c]{CelebA\\ (Feature: Male)\\Original FID: 166.3}}}
& H   & 4.345 (1.849)         & 174.8 (1.9) {\color{green!60!black}\ding{51}}         & 0.016 (0.0001)   \\
  & \shat   & 0.181 (0.209)         & 176.4 (3.5) {\color{green!60!black}\ding{51}}         & 0.023 (0.0002) $^*$ \\
  & \unoh                   & 0.499 (0.071)& \textbf{173.4} (1.8) {\color{green!60!black}\ding{51}}& 0.178 (0.0010) \\
  & \unosh                  & \textbf{0.390} (0.218)         & 175.2 (2.3) {\color{green!60!black}\ding{51}}         & 0.134 (0.0696)\\
\bottomrule
\end{tabular}
\end{table}
\endgroup

%%%%%%%%%%%%%%%%%%%%%%%%%%%%%%%%%%%%%%%%%%%%%%%%%%%%%%%%%%%%%%%%%%%%%%%%%%%%%%

\section{Orthogonalization in a linear regression model}\label{app:lin}
To understand the effect of the orthogonalization term in the loss function \eqref{eq:loss-uno} 
%, which we recall here 
%\begin{align}
%    \mathcal{L}_{\rm UNO} = \frac{1}{|\mathcal D_r|}\sum_{x\in\mathcal D_r}\mathcal{L}(\mathcal{M}_\theta, x) +  \beta_o \left(\frac{\gr \cdot \gf}{\|\gr\|\|\gf\|}\right)^2, 
%    \label{eq:loss-uno_app}
%\end{align}
let us consider linearly related input ($x$) and output ($y$) data
\begin{align}
y = W^\star x + \zeta,
\end{align}
with $\zeta \sim{\mathcal{N}}(0,\sigma_l^2)$ for $x\in \mathbb{R}^d$, $y\in \mathbb{R}$ and $W\in \mathbb{R}^{1\times d}$. We consider two datasets, a retain dataset and a forget dataset, with samples
\begin{align}
x_r &\sim {\mathcal{N}}(\mu_r,\sigma_r^2)\\
x_f &\sim {\mathcal{N}}(\mu_f,\sigma_f^2).
\end{align}
Drawing $N_r$ and $N_f$ samples we construct the data matrices $X_r\in \mathbb{R}^{d\times N_r}$ and $X_f\in \mathbb{R}^{d\times N_f}$, and the combined set $X=[X_r \, X_f] \in \mathbb{R}^{d\times N}$ with $N=N_r+N_f$, with corresponding $Y_{r,f}\in \mathbb{R}^{1\times N_{r,f}}$ and  $Y\in \mathbb{R}^{1\times N}$.

We analyze now how a model, initially trained on the whole dataset $\{X,Y\}$ with linear regression, changes during a single gradient descent step, and will find that the orthogonality term induces a gradient descent along the direction of largest variance of the retain data $X_r$ and a gradient ascent step along the direction of largest variance of the forget data $X_f$. 

The cost function \eqref{eq:loss-uno} for the linear model, as a function of the model parameters $W$, is written as
\begin{align}
    \mathcal{L}_{\rm UNO}(W) = \|Y_r-W X_r\|^2  +  \beta_o \| \left[ \nabla \|Y_r-W X_r\|^2\right]^\top   \left[\nabla\|Y_f-W X_f\|^2\right] \|^2. 
    \label{eq:loss-lin_app}
\end{align}
The orthogonality term is readily evaluated as
\begin{align}
    & \| \left[ \nabla \|Y_r-W X_r\|^2\right]^\top   \left[\nabla\|Y_f-W X_f\|^2\right] \|^2 \nonumber \\
    & \qquad = 16 \| \left[ X_rY_r^\top  - X_rX_r^\top W^\top  \right]^\top  \left[  X_fY_f^\top  - X_fX_f^\top W^\top \right] \|^2.  
\end{align}

A gradient descent step starting from the model obtained from the whole dataset $X$ is given by
\begin{align}
W_1 = W_0 - \eta \nabla  \mathcal{L}_{\rm UNO}(W_0),
\end{align}
where 
\begin{align}
W_0 = YX^\top \left(XX^\top  \right)^{-1}
\end{align}
is the least-square solution for the full dataset. To simplify expressions we introduce the covariance matrices 
\begin{align}
\Phi_{r,f} &= X_{r,f}X_{r,f}^\top \in \mathbb{R}^{d\times d},
\end{align}
and the mismatch
\begin{align}
E_{r,f} &= Y_{r,f}-W_0X_{r,f}\in \mathbb{R}^{1\times N_{r,f}},
\end{align}
and form
\begin{align}
\theta_{r,f} &= \nabla\left[ \| Y_{r,f}-WX_{r,f}\|^2 \right] = -X_{r,f}E_{r,f}^\top  \in \mathbb{R}^{d\times 1},
\label{eq:theta_app}
\end{align}
which we readily identify as the gradients of the standard unregularized loss function for the linear model restricted to the retain and forget datasets, respectively. Note that $\theta_{r} = \gr$ and $\theta_f=\gf$ (cf. \eqref{eq:gf}--\eqref{eq:gr}). Since $XY^\top =X_rY_r^\top +X_fY_f^\top $ and $ XX^\top W_0^\top = (X_rX_r^\top +X_fX_f^\top )W_0^\top $, we have $\theta_r = -\theta_f$. 

Introducing for simplicity of exposition $\beta=64\eta \beta_0$, we obtain
\begin{align}
W_1 = W_0 - \left[ \eta I + \beta \|\theta_r\|^2 \left[\Phi_r - \Phi_f \right] \right] \theta_r,
\end{align}
which we can write as a two-step update
\begin{align}
W_{\tfrac12} &= W_0 - \left[\eta I + \beta \|\theta_r\|^2 \Phi_r\right] \theta_r \\
W_1 &= W_{\tfrac12} + \beta \|\theta_r\|^2 \Phi_f \theta_r.
\end{align}
Hence, for the linear model, the orthogonality term leads to a gradient descent step predominantly in the dominant eigendirection of the covariance matrix $\Phi_r$ of the retain dataset $X_r$, followed by a gradient ascent step predominantly in the dominant eigendirection of the covariance matrix $\Phi_f$ of the forget dataset $X_f$.

\newpage

\end{document}